\documentclass[letterpaper, 10 pt, conference]{ieeeconf}  

\IEEEoverridecommandlockouts                              

\usepackage{amsmath} 
\usepackage{amssymb}  
\usepackage{adjustbox}
\usepackage{bm}  
\usepackage{graphics} 
\usepackage{epsfig} 
\usepackage{wrapfig}
\usepackage{subcaption}
\usepackage[font=small,labelfont=small]{caption}
\usepackage{hyperref}
\usepackage{color}
\usepackage[table]{xcolor}
\usepackage[ruled, linesnumbered, vlined]{algorithm2e}
\usepackage{sidecap}
\usepackage{tikz}
\usetikzlibrary{arrows.meta}
\usepackage[noadjust]{cite}  
\usepackage{cuted}
\usepackage{multirow}
\usepackage{booktabs}
\usepackage{amsthm}  
\usepackage{soul}

\newcommand{\trsp}{\mathsf{T}}  

\newcommand{\euclideanspace}{\mathbb{R}}
\newcommand{\manifold}{\mathcal{M}}
\newcommand{\tangentspace}[1]{\mathcal{T}_{#1}\mathcal{M}}

\newcommand{\norm}[2]{\| #2\|_{#1}}  

\newcommand{\expmapblank}[1]{\text{Exp}_{#1}}  
\newcommand{\logmapblank}[1]{\text{Log}_{#1}}  
\newcommand{\expmap}[2]{\expmapblank{#1}\left(#2\right)}  
\newcommand{\logmap}[2]{\logmapblank{#1}\left(#2\right)}  
\newcommand{\prltrspblank}[2]{\Gamma_{#1 \rightarrow #2}}  
\newcommand{\prltrsp}[3]{\prltrspblank{#1}{#2}(#3)}  

\newcommand{\sphere}[1]{\mathcal{S}^{#1}}

\definecolor{lightred}{rgb}{0.8, 0.22, 0.29}
\definecolor{turquoise}{rgb}{0.25, 0.89, 0.82}
\definecolor{mediumorchid}{rgb}{0.73, 0.33, 0.83}
\definecolor{darkorange}{rgb}{1.0, 0.65, 0.0}
\definecolor{olive}{rgb}{0.502, 0.502, 0.0}
\definecolor{royalblue}{rgb}{0.255, 0.412, 0.882}
\definecolor{dodgerblue}{rgb}{0.118, 0.565, 1.0}
\definecolor{orange}{rgb}{1.0, 0.647, 0.0}
\DeclareRobustCommand{\blackline}{\raisebox{2pt}{\tikz{\draw[black,solid,line width = 1.1pt](0,0) -- (4mm,0);}}}
\DeclareRobustCommand{\dodgerblueline}{\raisebox{2pt}{\tikz{\draw[dodgerblue,solid,line width = 1.1pt](0,0) -- (4mm,0);}}}
\DeclareRobustCommand{\royalblueline}{\raisebox{2pt}{\tikz{\draw[royalblue,solid,line width = 1.1pt](0,0) -- (4mm,0);}}}
\DeclareRobustCommand{\orangeline}{\raisebox{2pt}{\tikz{\draw[orange,solid,line width = 1.1pt](0,0) -- (4mm,0);}}}
\DeclareRobustCommand{\oliveline}{\raisebox{2pt}{\tikz{\draw[olive,solid,line width = 1.1pt](0,0) -- (4mm,0);}}}
\DeclareRobustCommand{\olivecircle}{\tikz{ \filldraw[color=white, fill=olive, thick](0,0) circle (.075);}}
\DeclareRobustCommand{\bluecircle}{\tikz{ \filldraw[color=white, fill=blue, thick](0,0) circle (.075);}}

\graphicspath{{Figures/}}

\title{\LARGE \bf Riemannian Flow Matching Policy for Robot Motion Learning}

\author{Max Braun$^{1}$, No\'emie Jaquier$^{1}$, Leonel Rozo$^{2}$, and Tamim Asfour$^{1}$
\thanks{This work was supported by the Carl Zeiss Foundation under the project JuBot and the European Union's Horizon Europe Framework Programme under grant agreement No 101070596 (euROBIN).	}
\thanks{$^{1}$Institute for Anthropomatics and Robotics, Karlsruhe Institute of Technology, Karlsruhe, Germany. 
\href{mailto:noemie.jaquier@kit.edu}{\textrm{noemie.jaquier@kit.edu}}, 
\href{mailto:asfour@kit.edu}{\textrm{asfour@kit.edu}}
}
\thanks{$^{2}$Bosch Center for Artificial Intelligence. Renningen, Germany. \href{mailto:leonel.rozo@de.bosch.com}{\textrm{leonel.rozo@de.bosch.com}}}
}

\begin{document}

\makeatletter

\makeatother
\maketitle

\thispagestyle{empty}
\pagestyle{empty}

\begin{abstract}
We introduce Riemannian Flow Matching Policies (RFMP), a novel model for learning and synthesizing robot sensorimotor policies. 
RFMP leverages the efficient training and inference capabilities of flow matching methods.
By design, RFMP inherits the strengths of flow matching: the ability to encode high-dimensional multimodal distributions, commonly encountered in robotic tasks, and a very simple and fast inference process.
We demonstrate the applicability of RFMP to both state-based and vision-conditioned robot motion policies. 
Notably, as the robot state resides on a Riemannian manifold, RFMP inherently incorporates geometric awareness, which is crucial for realistic robotic tasks.
To evaluate RFMP, we conduct two proof-of-concept experiments, comparing its performance against Diffusion Policies.
Although both approaches successfully learn the considered tasks, our results show that RFMP provides smoother action trajectories with significantly lower inference times.

\end{abstract}

\section{INTRODUCTION}
The problem of learning, synthesizing and adapting robot motions in unstructured environments has been recently disrupted by the rise of deep generative models.
These models enable a robot to learn elaborated skills that may display high-dimensional multimodal action distributions. 
They can also be interfaced with deep multimodal perception networks, thus allowing a robot to learn sensorimotor policies.
These models have been leveraged in both imitation and reinforcement settings~\cite{Janner22:DiffPlanning,Reuss23:GoalCondDiff,Wang23:DiffPolORL,Cheng23:DiffPolicy}, where diffusion processes~\cite{luo22:DiffModels,Yang23:DiffModelsSurvey} have recently shown promising results in a plethora of real robotic tasks.
Nevertheless, this type of models are characterized by expensive inference methods as they often require to solve a stochastic differential equation, which might hinder their use in some robotic settings~\cite{luo22:DiffModels}, e.g., for reactive motion policies.
Moreover, when learning diffusion models on Riemannian manifolds, the computation of the score function of the diffusion process is not as simple as in the Euclidean case~\cite{Huang22:RiemannianDM}, and the inference process incurs increasing computational complexity.

\begin{figure}[th]
    \centering
		\includegraphics[trim={4cm 5cm 4cm 4cm}, clip, width=.23\textwidth]{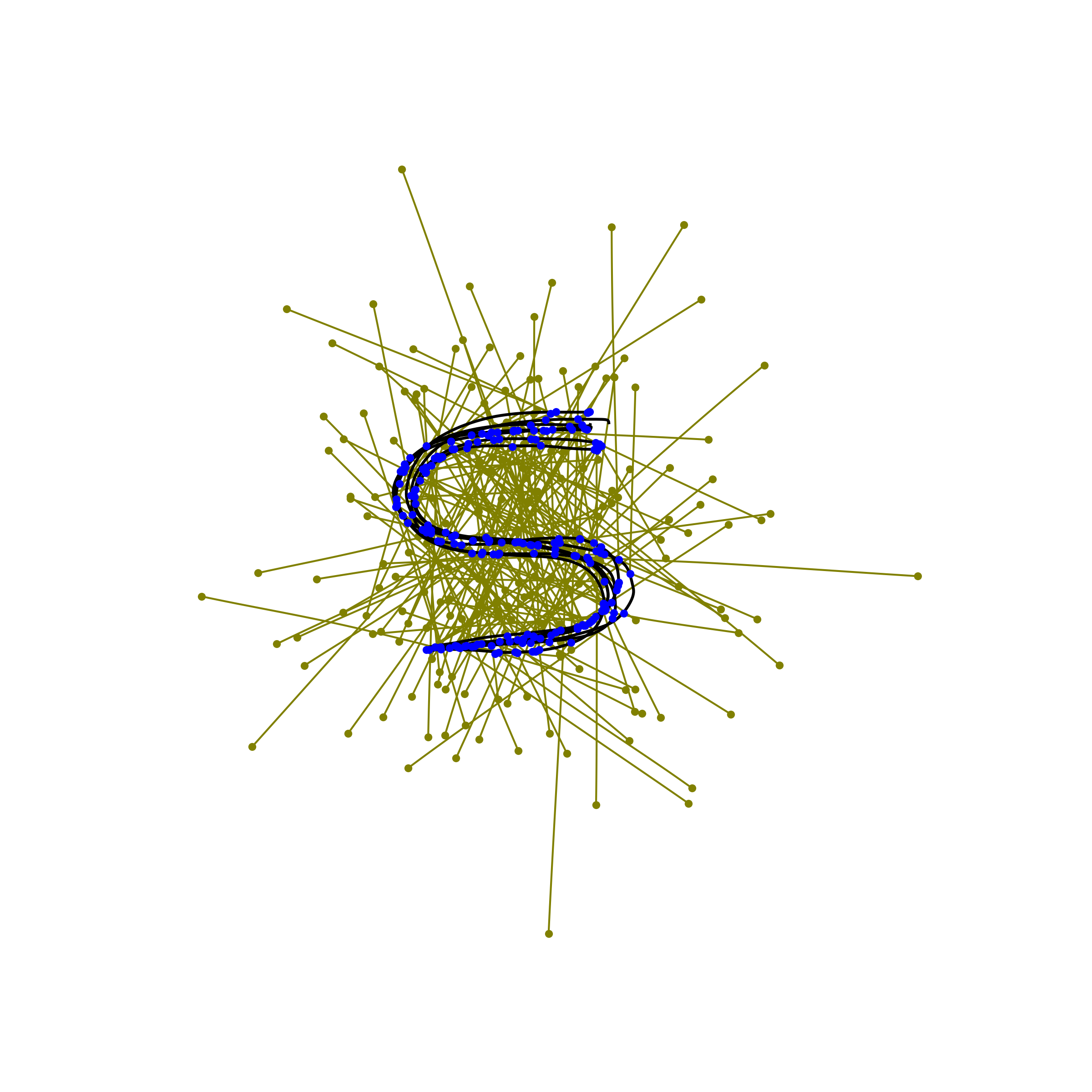}
        \includegraphics[trim={4cm 3cm 4cm 5cm}, clip, width=.23\textwidth]{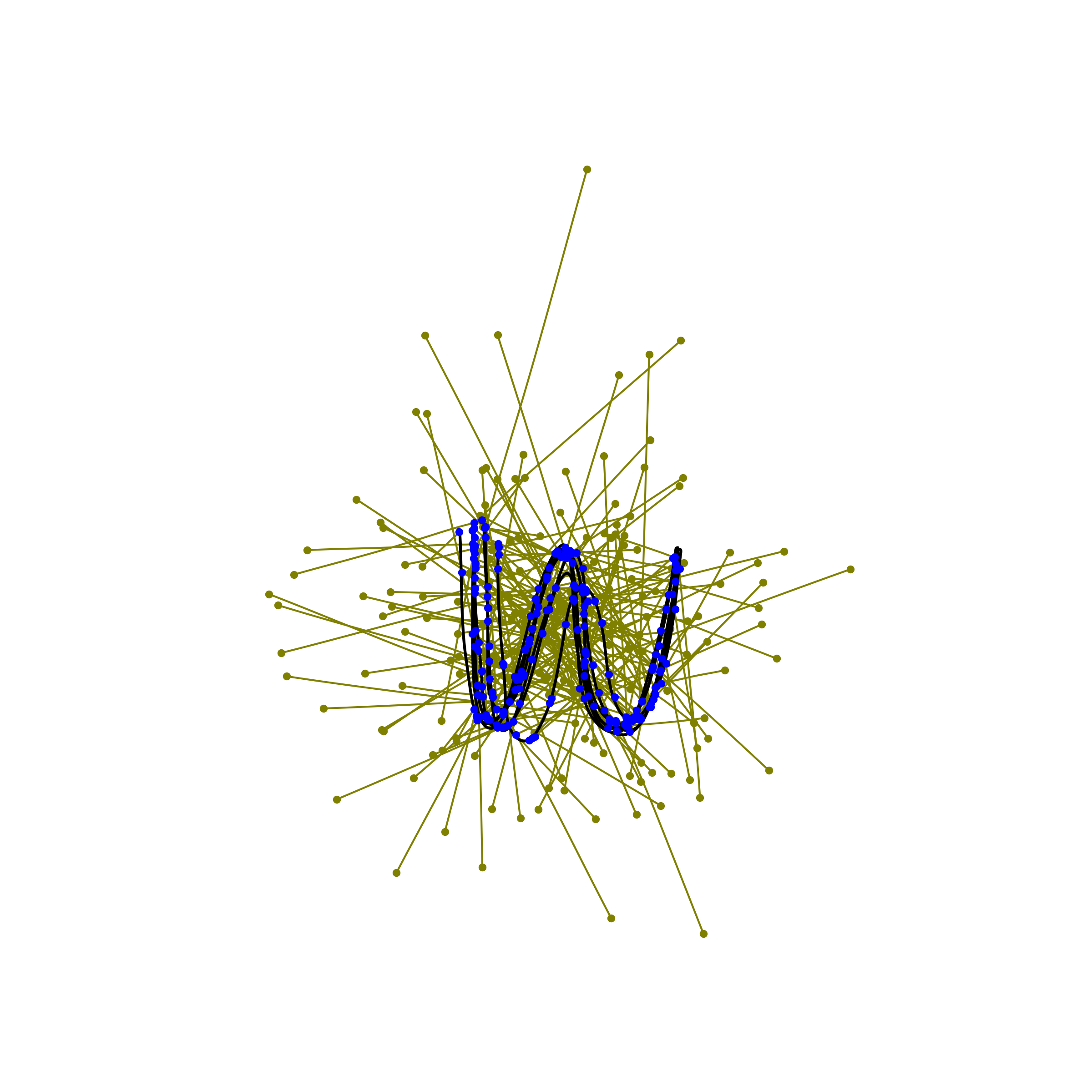}
		\includegraphics[trim={4cm 4cm 4cm 4cm}, clip, width=.23\textwidth]{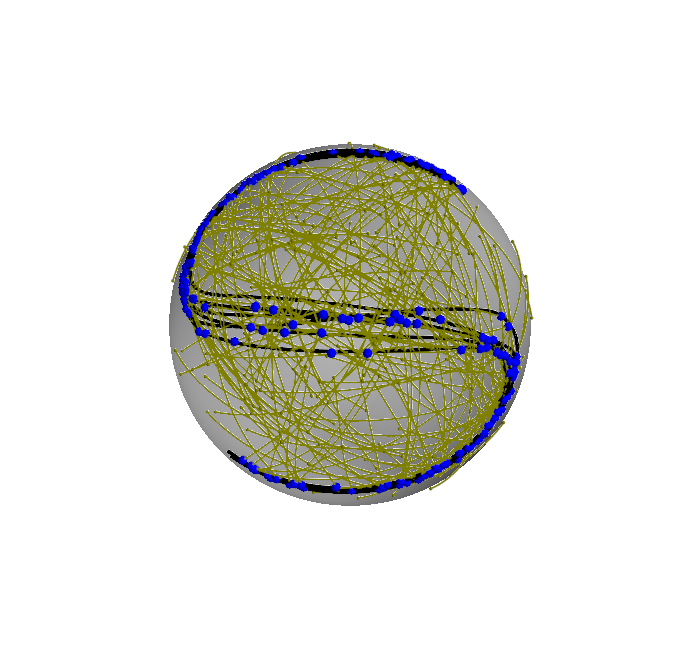}
        \includegraphics[trim={4cm 4cm 4cm 4cm}, clip, width=.23\textwidth]{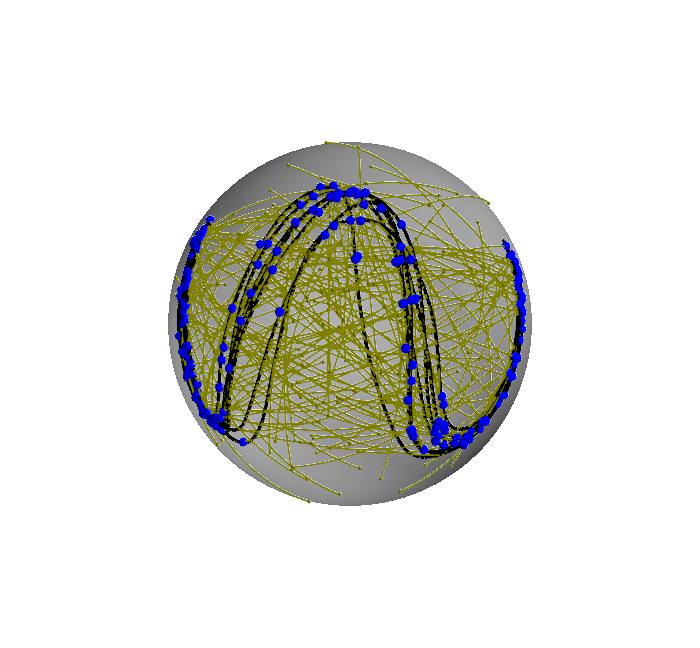}
	\caption{Learned RFMP flows (\oliveline) from the base distribution (\olivecircle) to the LASA datasets $\mathsf{S}$ and $\mathsf{W}$ (\bluecircle) on both $\mathbb{R}^2$ (top) and $\sphere{2}$ (bottom). The flow is conditioned on random observations $\bm{o}$ from the training dataset (\blackline). }
	\label{Fig:RMFPsamples}
    \vspace{-0.5cm}
\end{figure}

In contrast to diffusion models, flow matching (FM)~\cite{Lipman23:FlowMatching} takes a different approach.
Intuitively, FM defines a series of small transformations (flows) that can smoothly move samples from a base distribution towards the target data points (i.e., the demonstration dataset).
Each flow is represented by simple function that takes a base distribution sample and pushes it slightly in a specific direction. 
By chaining these small flows together, FM gradually transforms the prior distribution into the target (demonstrations) distribution.
The beauty of FM lies in its simplicity, as these flow function is much easier to train and evaluate compared to solving complex stochastic differential equations as in diffusion models.
Motivated by the recent efficacy of flow matching (FM) methods~\cite{Lipman23:FlowMatching} across various machine learning domains~\cite{Davtyan23:FMvideo,Liu24:FMspeech,Bose24:FMproteinSE3}, we propose to learn sensorimotor robot skills via a Riemannian Flow Matching Policy (RFMP).
RFMP capitalizes on the easy training and fast inference of FM methods to learn and synthesize robot movements represented by end-effector pose trajectories. 
Our main contributions are twofold: (1) we pioneer the application of FM methods within sensorimotor robot policies learning, and (2) we empirically validate their effectiveness on the established benchmark LASA dataset~\cite{Lemme2015:LasaDataset}.

\textbf{Related work}:
The literature on robot policy learning is vast, and therefore we focus on approaches that design policies based on flow-based generative models.
Normalizing flows~\cite{Papamakarios21:NormalizingFlows} are arguably the first models to be broadly adapted as robot policy representations. 
The most common approach was to employ them as diffeomorphisms for learning stable dynamical systems~\cite{Rana2020:EuclideanizingFlows,Khader21:StableNF,Urain20:ImitationFlow}, with extensions to Lie groups~\cite{Urain22:StableSE3}, and Riemannian manifolds~\cite{Zhang23:RiemannianStableDS}.
A shortcoming of normalizing flows is their slow training due to the integration of the associated ODE.
More recently, diffusion models have dominated the robot learning scene due to their more stable training and their ability to learn complex data distributions more accurately~\cite{Yang23:DiffModelsSurvey}.
They have been primarily employed to learn motion planners~\cite{Janner22:DiffPlanning} and complex control policies~\cite{Cheng23:DiffPolicy,Wang23:DiffPolORL,Reuss23:GoalCondDiff}.
In contrast to the aforementioned works, our work leverages flow matching~\cite{Lipman23:FlowMatching} to model robot motion policies.
This choice stems from the inherent advantages of FM: it avoids the complex training procedures of normalizing flows and the computationally expensive inference of diffusion models.
Furthermore, our method also accounts for full-pose trajectories by leveraging the recently developed Riemannian extension of FM models presented in~\cite{Chen24:RiemannianFM}.

\section{BACKGROUND}
\label{sec:Background}
In this section, we provide a short background on Riemannian geometry, an overview of the general flow matching framework and its extension to Riemannian manifolds.

\subsection{Riemannian manifolds}
Let us imagine a flexible and smoothly-curved surface like a globe. 
A smooth manifold, mathematically denoted by $\manifold$, can be intuitively conceptualized as the globe surface.
Locally, a small patch of this manifold looks flat, therefore resembling the Euclidean space $\euclideanspace^d$~\cite{DoCarmo92:RiemannianGeometry, Lee18:RiemannianManifolds}.  
The smoothness of the manifold allows us to define directions and rates of change at each point, leading to tangent vectors in $\euclideanspace^d$.
The set of tangent vectors of all curves at $\bm{x} \in \manifold$ forms a vector space isomorphic to $\euclideanspace^d$, known as the \emph{tangent space} $\tangentspace{\bm{x}}$ of $\manifold$ at $\bm{x}$.
The collection of all such tangent spaces is called the \emph{tangent bundle} $\tangentspace{} = \bigcup_{\bm{x}\in\manifold} \left\{(\bm{x}, \bm{u}) | \bm{u} \in \tangentspace{\bm{x}} \right\}$
It is possible to endow a smooth manifold $\manifold$ with a \emph{Riemannian metric}, which is a family of inner products $g_{\bm{x}}: \tangentspace{\bm{x}} \times  \tangentspace{\bm{x}} \rightarrow \euclideanspace$ associated to each point $\bm{x} \in \manifold$.
A \emph{Riemannian manifold} $(\manifold, g)$ is a smooth manifold endowed with a Riemannian metric $g$, that is a family of inner products $g_{\bm{x}}: \tangentspace{\bm{x}} \times \tangentspace{\bm{x}} \rightarrow \euclideanspace$ associated to each point $\bm{x} \in \manifold$~\cite{Lee18:RiemannianManifolds}.

To operate with Riemannian manifolds, we can leverage their Euclidean tangent spaces and resort to mappings back and forth between $\tangentspace{\bm{x}}$ and $\manifold$, using the exponential and logarithmic maps.
Specifically, the exponential map $\expmap{\bm{x}}{\bm{u}}: \tangentspace{\bm{x}} \to \manifold$ maps a point $\bm{u}\in\tangentspace{\bm{x}}$ to a point $\bm{y}$ on the manifold, so that it lies on the geodesic starting at $\bm{x}$ in the direction $\bm{u}$ and such that the geodesic distance $d_{\manifold}(\bm{x}, \bm{y}) = d_{\euclideanspace, g_{\bm{x}}}(\bm{x}, \bm{u})$. 
The inverse operation is the logarithmic map $\logmap{\bm{x}}{\bm{y}}: \manifold \to \tangentspace{\bm{x}}$ .
Finally, the parallel transport $\prltrsp{\bm{x}}{\bm{y}}{\bm{u}}: \tangentspace{\bm{x}}\to\tangentspace{\bm{y}}$ describes how elements of $\manifold$ can be transported along curves on $\manifold$ while maintaining their intrinsic geometric properties. 
This allows us to operate elements lying on different tangent spaces.

\subsection{Flow Matching}
Flow matching~\cite{Lipman23:FlowMatching} is a simulation-free generative model that reshapes a simple base density $p_0 \in \mathbb{P}(\mathbb{R}^d)$ to a target (more complicated) distribution $p_1 \in \mathbb{P}(\mathbb{R}^d)$ via the push-forward of the prior $p_1 = \phi_{\sharp} p_0$, with $\phi$ denoting the \emph{flow} and $\sharp$ being the push-forward operator. 
To design this flow, we can define a vector field $u_t: [0,1] \times \mathbb{R}^d \rightarrow \mathbb{R}^d$ that represents the ODE, 
\begin{equation}
    \frac{d\phi_t(\bm{x})}{dt} = u_t(\phi_t(\bm{x})) \quad \text{with initial condition} \quad \phi_0(\bm{x}) = \bm{x} .
    \label{Eq:ODE}
\end{equation}
Loosely speaking, the vector field $u_t$ defines how a sample $\bm{x}_0 \sim p_0$ is transformed over time (from $t_0$ to $t=1$) to match a target sample from $\bm{x}_1 \sim p_1$.
At a density level, the vector field defines a probability density path $p_t : [0,1] \times \mathbb{R}^d \rightarrow \mathbb{R}^d$, i.e. an interpolation in probability space, which is characterized by the continuity equation~\cite{Lipman23:FlowMatching}. 

Assuming that both the probability path $p_t(\bm{x})$ and the corresponding vector field $u_t(\bm{x})$ are known, then one could regress a parametrized vector field $v_t(\cdot;\bm{\theta}): [0,1] \times \mathbb{R}^d \rightarrow \mathbb{R}^d$ to some target vector field $u_t$, which leads to the FM loss,
\begin{equation}
    \mathcal{L}_{\text{FM}}(\bm{\theta}) = \mathbb{E}_{t,p_t(\bm{x})} \norm{2}{v_t(\bm{x};\bm{\theta}) - u_t(\bm{x})}^2 , 
    \label{Eq:FMintractableloss}
\end{equation}
where $\bm{\theta}$ are the learnable parameters, $t\sim\mathcal{U}[0,1]$, and $\bm{x} \sim p_t(\bm{x})$. 
Unfortunately, the objective in~\eqref{Eq:FMintractableloss} is intractable since we actually do not have prior knowledge about $p_t$ and $u_t$. 
Lipman \emph{et al.}~\cite{Lipman23:FlowMatching} showed that by defining a conditional probability path $p_t(\bm{x}|\bm{z})$ (and consequently a conditional vector field $u_t(\bm{x}|\bm{z})$), it is possible to obtain a tractable conditional flow matching (CFM) loss,
\begin{equation}
    \mathcal{L}_{\text{CFM}}(\bm{\theta}) = \mathbb{E}_{t,q(\bm{z}),p_t(\bm{x}|\bm{z})} \norm{2}{v_t(\bm{x};\bm{\theta}) - u_t(\bm{x}|\bm{z})}^2 , 
    \label{Eq:FMloss}
\end{equation}
which happens to have identical gradients to the unconditional loss~\eqref{Eq:FMintractableloss} w.r.t $\bm{\theta}$.

Tong \emph{et al.}~\cite{Tong23:CFMoptimaltransport} showed that there exist several forms of CFM depending on how we design the prior $q(\bm{z})$, the conditional probability $p_t(\bm{x}|\bm{z})$, and the corresponding vector field $u(\bm{x}|\bm{z})$.
For example, by considering the conditioning variable as $\bm{z} = \bm{x}_1 \sim p_1$, and by defining $p_t(\bm{x}|\bm{z})$ and $u(\bm{x}|\bm{z})$ as follows,
\begin{align}
    p_t(\bm{x}|\bm{z}) &= \mathcal{N}\left(\bm{x}| t\bm{x}_1, (t\sigma-t+1)^2\right) , \\
    u_t(\bm{x}|\bm{z}) &= \frac{\bm{x}_1 - (1-\sigma)\bm{x}}{1-(1-\sigma)t} , 
    \label{Eq:LipmanCFM}
\end{align}
one recovers the Gaussian CFM of Lipman \emph{et al.}~\cite{Lipman23:FlowMatching}, which defines a probability path from a zero-mean normal distribution to a Gaussian distribution centered at $\bm{x}_1$, which is also the approach taken in this paper.
Finally, the inference process boils down to: (1) Get a sample from $p_0$ and; (2) Use the vector field $v_t(\bm{x};\bm{\theta})$ to solve the ODE~\eqref{Eq:ODE} using off-the-shelf solvers.

\textbf{The Riemannian case:}
In several robotic settings, the target data distribution may lie on a Riemannian manifold $\mathcal{M}$, as the desired movements for a robot's end-effector often include the orientation component. 
Therefore, the part of the robot state representation lie on either the $\mathcal{S}^3$ hypersphere  or the $\operatorname{SO}(3)$ group, depending on the specific parametrization used.
To properly handle this type of cases, Chen and Lipman~\cite{Chen24:RiemannianFM} recently extended CFM to Riemannian manifolds (RCFM).
Formally, this Riemannian formulation considers that $\bm{x} \in \mathcal{M}$, and therefore the vector field $u_t(\bm{x}) \in \mathcal{T}_{\bm{x}}\mathcal{M}$ (i.e., it evolves on the tangent bundle $\mathcal{TM}$) generates a probability density path $p_t \in \mathbb{P}(\mathcal{M})$.
As stated previously, a Riemannian manifold $\mathcal{M}$ is endowed with a Riemannian metric $g_{\bm{x}}$, which implies that the CFM loss~\eqref{Eq:FMloss} is now computed w.r.t such a metric, as follows,
\begin{equation}
    \mathcal{L}_{\text{RCFM}}(\bm{\theta}) = \mathbb{E}_{t,q(\bm{z}),p_t(\bm{x}|\bm{z})} \norm{g_{\bm{x}}}{v_t(\bm{x};\bm{\theta}) - u_t(\bm{x}|\bm{z})}^2 .
    \label{Eq:RCFMloss}
\end{equation}

As in the Euclidean case, we need to design the vector field and choose the base distribution. 
Following~\cite{Chen24:RiemannianFM,Bose24:FMproteinSE3}, the most straightforward strategy is to exploit geodesic paths to design the flow $\phi$, i.e., we use the shortest path to connect $\bm{x}_0$ and $\bm{x}_1$.
Importantly, for many known geometries such as the $\mathcal{S}^d$ hypersphere, the $\operatorname{SO}(3)$ group, or the manifold of symmetric positive definite matrices $\mathcal{S}_{++}^d$, to mention a few, we have closed-form geodesics.
Specifically, the geodesic flow connecting $\bm{x}_0$ and $\bm{x}_1$ is given by,
\begin{equation}
    \bm{x}_t = \expmap{\bm{x}_0}{t\,\logmap{\bm{x}_0}{\bm{x}_1}} , \quad t \in [0,1] .
    \label{Eq:GeodesicInterpolant}
\end{equation}
We now can design the vector field $u_t(\bm{x}|\bm{z})$ by leveraging the ODE associated with the conditional flow $d\phi_t(\bm{x})/dt~=~\Dot{\bm{x}}_t~=~u_t(\bm{x}|\bm{z})$.
This means that computing the vector field $u_t(\bm{x}|\bm{z})$ corresponds to compute the time derivative of~\eqref{Eq:GeodesicInterpolant}. 
Finally, the choice of the base distribution $p_0$ generally depends on the problem at hand. 
One could directly define the base density as a uniform distribution over $\mathcal{M}$ as in~\cite{Chen24:RiemannianFM,Bose24:FMproteinSE3}, but it is also possible to use Riemannian or wrapped Gaussian distributions, as we do next.

\section{The Riemannian Flow Matching Policy}
\label{sec:FMP}
Given a set of trajectories $\mathcal{D} =\{\bm{o}_n, \bm{a}_n\}_{n=1}^N$, where $\bm{o}$ denotes the observation and $\bm{a}$ represents the corresponding action, our goal is to leverage the CFM framework to learn a Riemannian flow matching policy (RFMP) $\pi_{\bm{\theta}}(\bm{a}|\bm{o})$. This policy aims to generate action sequences that adhere to the target (expert) distribution $\pi_e$.
Note that, in the general case, we assume that both $\bm{o},\bm{a} \in \manifold$.
We hereinafter explain how we leverage CFM to model, train, and use such a policy.

\begin{figure*}[t]
    \centering
    \captionsetup{type=figure}
	\begin{subfigure}[b]{\linewidth}
        \centering
		\includegraphics[trim={1cm 2cm 1cm 3cm}, clip, width=.22\textwidth]{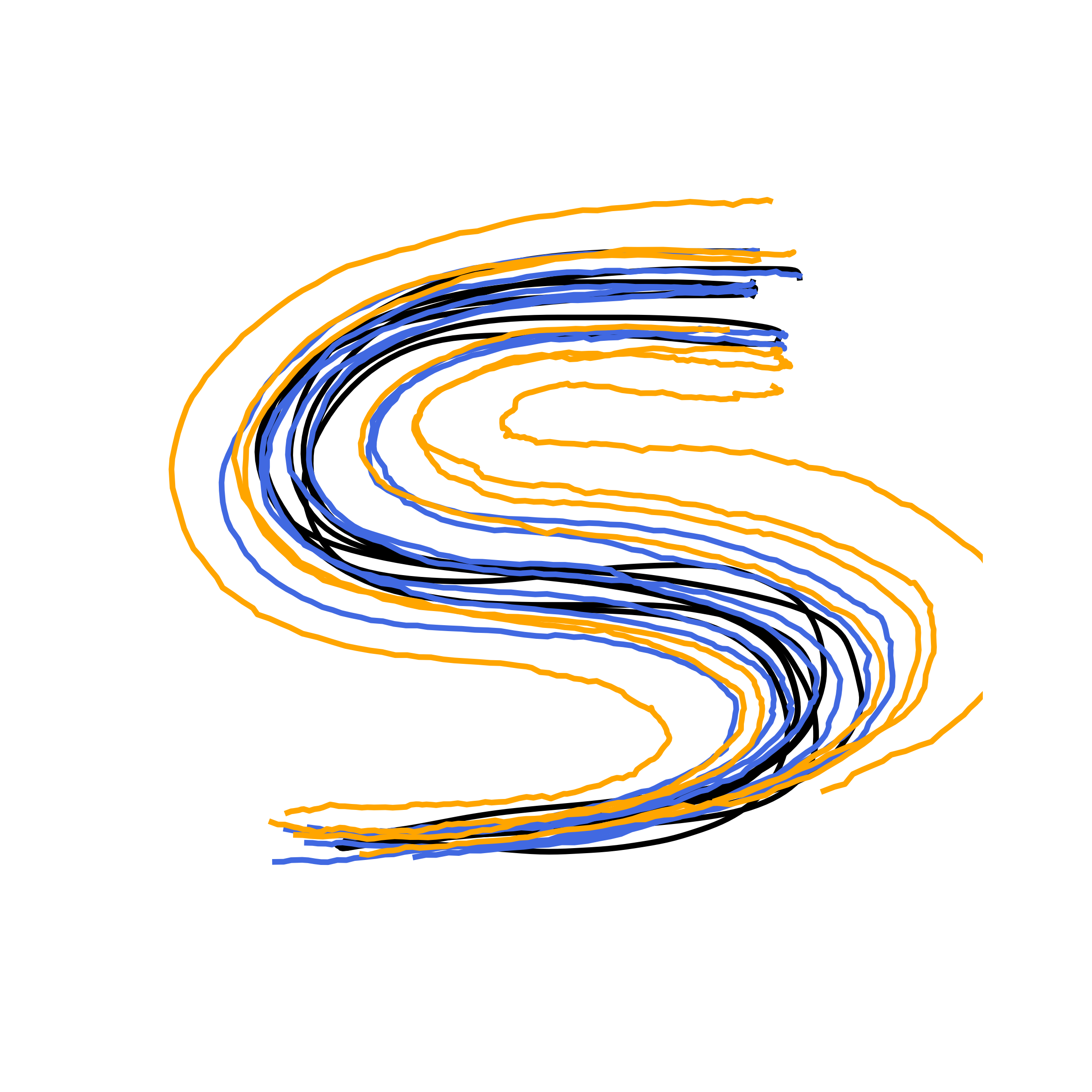}
        \includegraphics[trim={4cm 4cm 4cm 4cm}, clip, width=.22\textwidth]{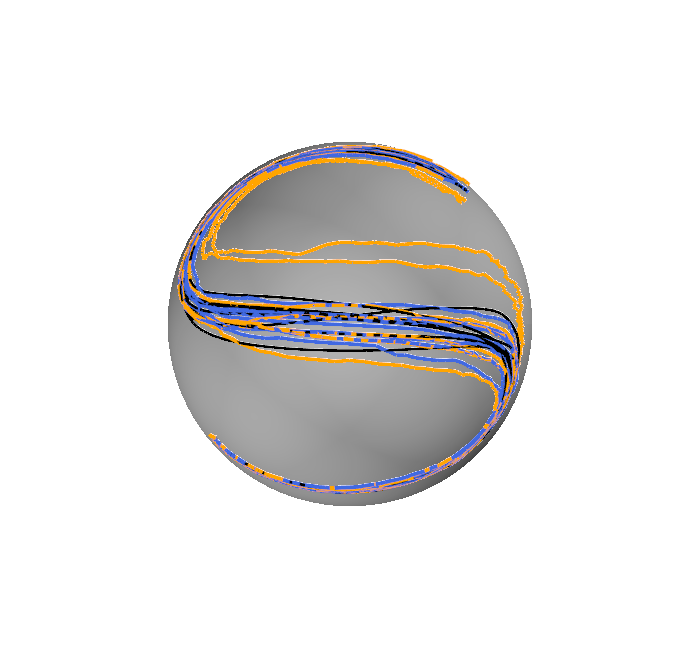}
        \includegraphics[trim={4cm 4cm 4cm 4cm}, clip, width=.22\textwidth]{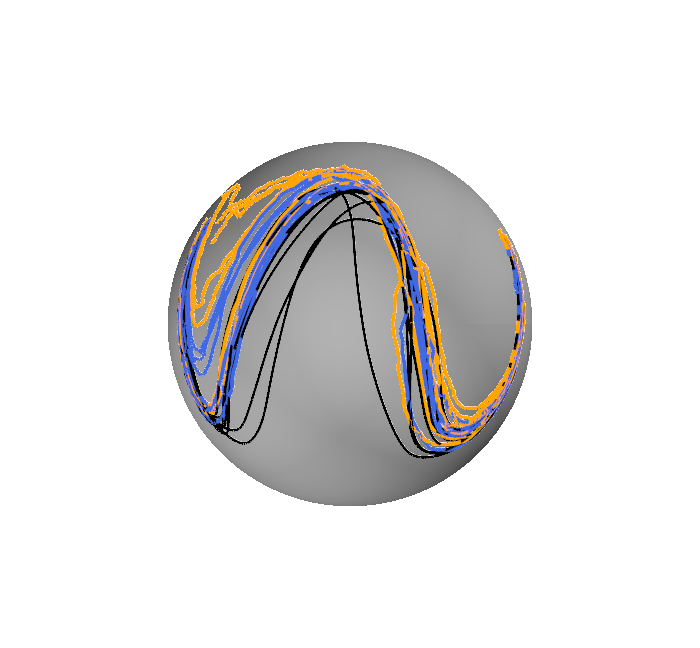}
        \includegraphics[trim={4cm 4cm 4cm 4cm}, clip, width=.22\textwidth]{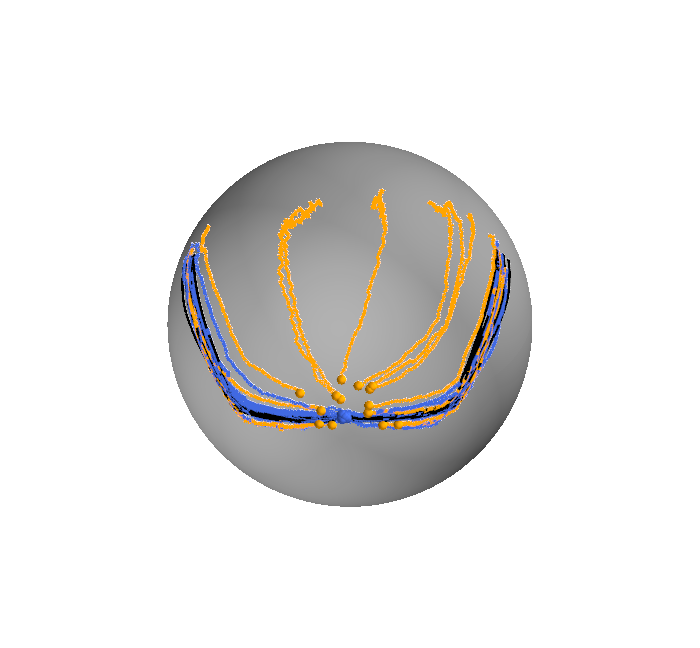}
		\caption{RFMP trajectories.} 
		\label{subFig:RFMP-trajectories}
	\end{subfigure}
	\begin{subfigure}[b]{\linewidth}
        \centering
		\includegraphics[trim={1cm 2cm 1cm 3cm}, clip, width=.22\textwidth]{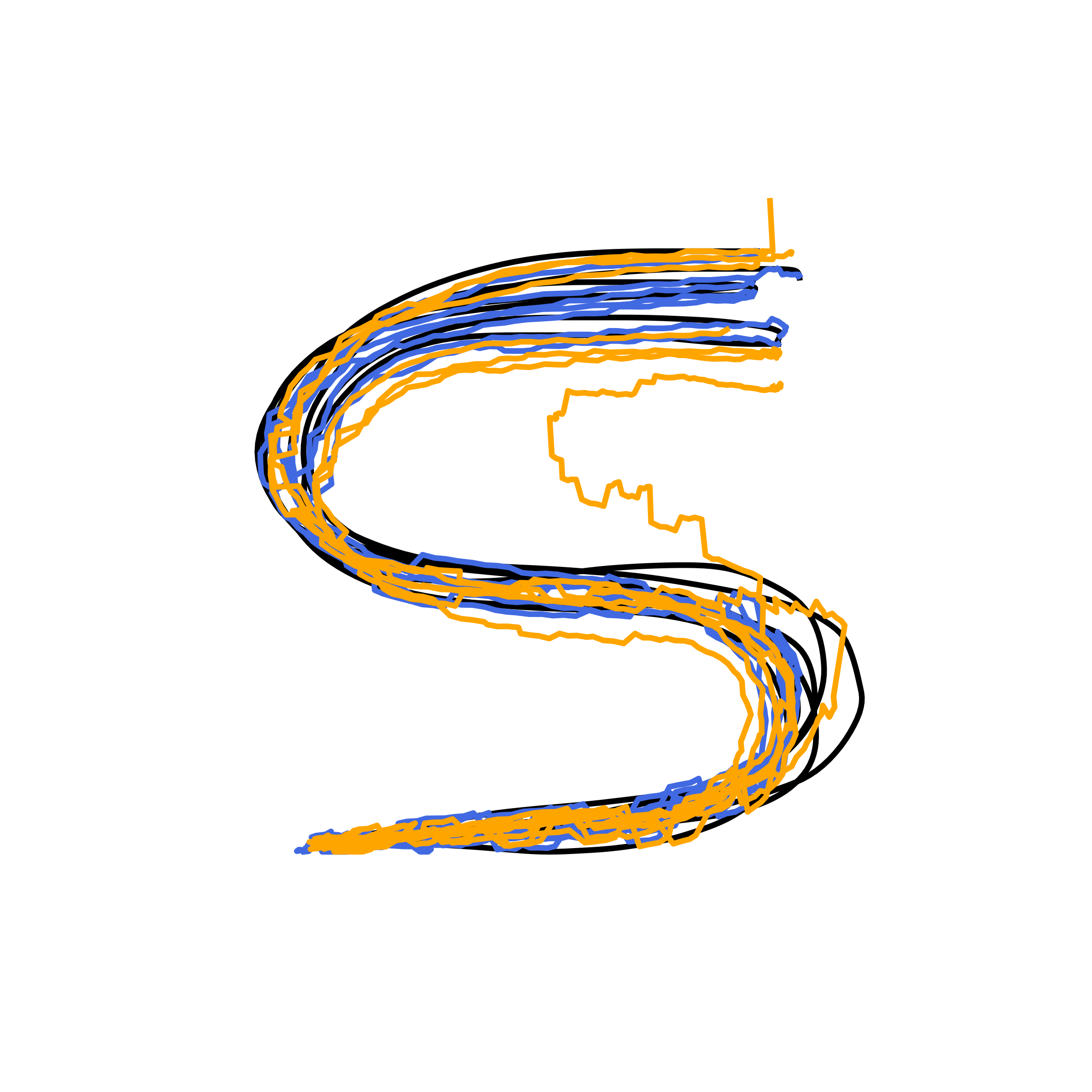}
        \includegraphics[trim={4cm 4cm 4cm 4cm}, clip, width=.22\textwidth]{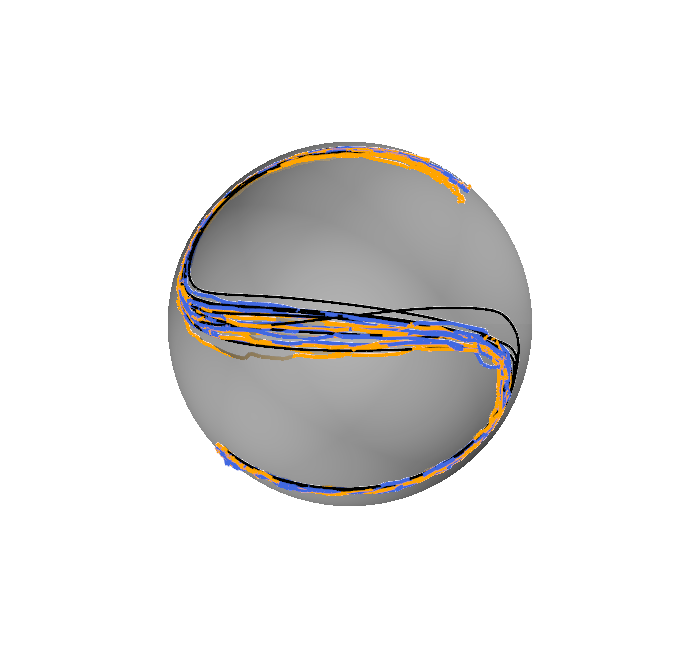}
        \includegraphics[trim={4cm 4cm 4cm 4cm}, clip, width=.22\textwidth]{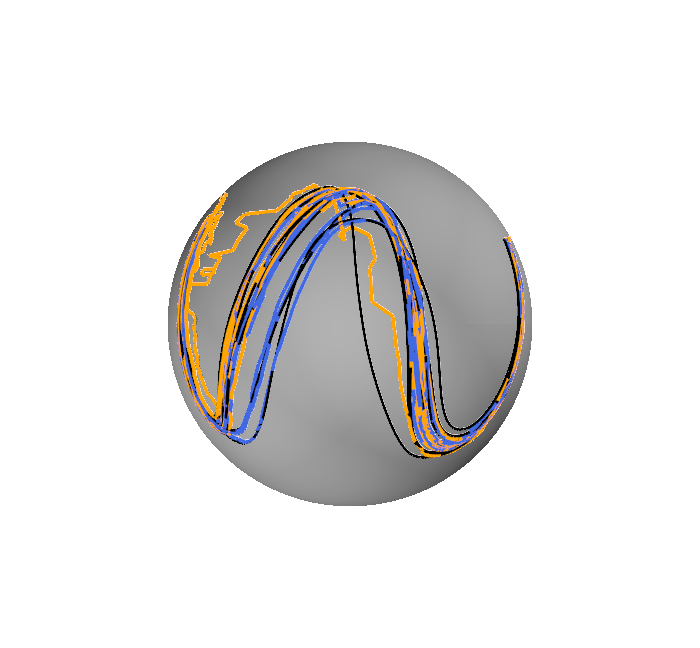}
        \includegraphics[trim={4cm 4cm 4cm 4cm}, clip, width=.22\textwidth]{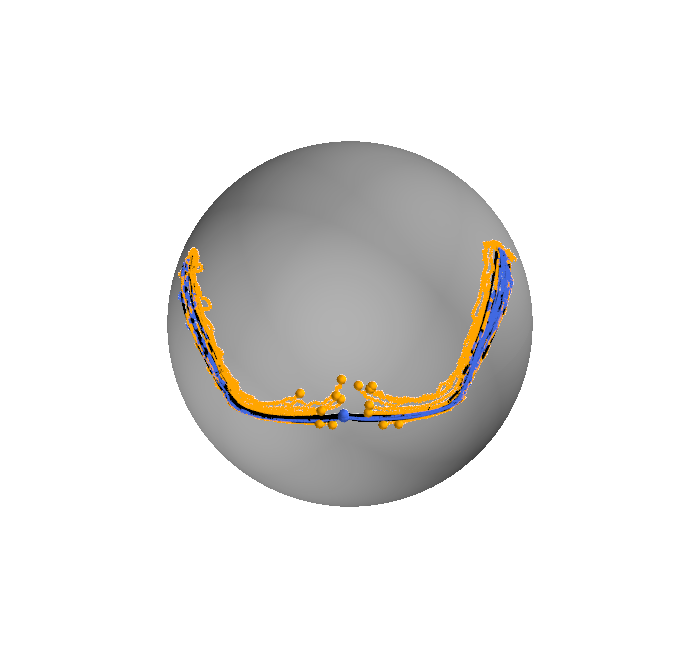}
		\caption{DP trajectories. } 
		\label{subFig:DP-trajectories}
	\end{subfigure}
	\caption{Demonstrations (\blackline) and learned trajectories on the LASA datasets $\mathsf{S}$ in $\euclideanspace^2$ (\emph{left}), on the LASA datasets $\mathsf{S}$, $\mathsf{W}$ projected on $\sphere{2}$ (\emph{middle-left, middle-right}), and on a multimodal dataset made of mirrored datasets of the letter $\mathsf{L}$ projected on $\sphere{2}$ (\emph{right}). Reproductions start at the same initial observations as the demonstrations (\royalblueline), or from randomly-sampled observations in the demonstration dataset neighborhood (\orangeline). Trajectory starts are depicted by dots in the multimodal case.}
	\label{Fig:RMFP-trajectories}
    \vspace{-0.3cm}
\end{figure*}

\subsection{RFMP training} Firstly, we adapt RCFM to visuomotor policies by simply conditioning the parametrized vector field on the observation vector $\bm{o}_t$, that is $v_t(\bm{a}|\bm{o})$.
Secondly, inspired by the diffusion policies framework~\cite{Cheng23:DiffPolicy}, we employ a receding horizon to achieve temporal consistency and smoothness on the predicted actions.
This means that our predicted action horizon vector is constructed as $\bm{a} = [\bm{a}_\tau, \bm{a}_{\tau+1}, \ldots, \bm{a}_{\tau+T_a} ]$ for a $T_a$-steps prediction horizon.
This implies that all samples $\bm{a}_1$ drawn from the target distribution have the form of the action horizon vector $\bm{a}$. 
Moreover, the samples $\bm{a}_0$ from the base distribution are constructed as $\bm{a}_0=[\bm{a}_{p_0}, \ldots, \bm{a}_{p_0}]$ with $\bm{a}_{p_0}\sim p_0$.
Nevertheless, instead of defining a similar receding horizon for the observations, we randomly sample only $T_o$ observation vectors from the training dataset to construct the conditioning variable $\bm{o}$. 
To do so, we follow the sampling strategy proposed in~\cite{Davtyan23:FMvideo}, which uses: (1) A \emph{reference observation} $\bm{o}_{\tau-1}$ ; (2) A \emph{context observation} $\bm{o}_{c}$ with the index $c$ uniformly sampled from $\{1, \ldots, \tau-2\}$; and (3) The distance $\tau-c$ between the prediction and the context observation. 
The combination of a reference and a context observation overcomes the fact that a single observation carries very little information and provides additional information about the direction of the motion.
Therefore, the observation vector is defined as $\bm{o} = [\bm{o}_{\tau-1}, \bm{o}_{c}, \tau-c ]$. 
The aforementioned strategy leads to the following RFMP loss,
\begin{equation}
    \mathcal{L}_{\text{RFMP}}(\bm{\theta}) = \mathbb{E}_{t, q(\bm{a}_1), p_t(\bm{a}|\bm{a}_1)} \norm{g_{\bm{a}}}{v_t(\bm{a} | \bm{o};\bm{\theta}) - u_t(\bm{a}|\bm{a}_1)}^2 .
    \label{Eq:RFMPloss}
\end{equation}
Algorithm~\ref{Alg:RFMP} summarizes the training procedure of RFMP.

\begin{algorithm}[t]
	\caption{Riemannian Flow Matching Policy}
	\label{Algo:UnivariateGeodesicReg}
	\small
	\SetAlgoLined
	\DontPrintSemicolon
	\SetKwInOut{Input}{Input}
	\SetKwInOut{Output}{Output}
	\Input{Initial parameters $\bm{\theta}$, base and target distributions $q(\bm{a}_1), p(\bm{a}_1)$.}
	\Output{Regressed vector field parameters $\bm{\theta}$.}
	\While{termination condition}{
		Sample time step $t \sim \mathcal{U}$. \;
		Sample training example $\bm{a}_1 \sim p$, and noise $\bm{a}_0 \sim q$. \;
		Sample observation vector $\bm{o}$. \;
        Compute target vector field $\Dot{\bm{a}}_t=u_t(\bm{a}|\bm{a}_1)$ based on the geodesic flow~\eqref{Eq:GeodesicInterpolant}. \;
        Evaluate $\mathcal{L}_{\text{RFMP}}(\bm{\theta})$~\eqref{Eq:RFMPloss}. \;
        Update parameters $\bm{\theta}$. 
	}
 \label{Alg:RFMP}
\end{algorithm}

\subsection{RFMP inference} Once our RFMP is trained, the inference process, which corresponds to querying our policy $\pi_{\bm{\theta}}(\bm{a}|\bm{o})$, is carried out as follows: (1) Draw a sample $\bm{a}_0 \sim q$; (2) Employ an off-the-shelf ODE solver to integrate the learned vector field $v_t(\bm{a} | \bm{o};\bm{\theta})$ along the time interval $[0,1]$; (3) Execute only the first $T_e$ actions [$\bm{a}_\tau, \bm{a}_{\tau+1}, \ldots,  \bm{a}_{\tau+T_e}$] with $T_e<T_a$, from the whole predicted action horizon $\bm{a}$.
Note that the ODE solver queries the learned vector field $v_t(\bm{a} | \bm{o};\bm{\theta})$ using the observation vector $\bm{o} = [\bm{o}_{\tau-1}, \bm{o}_{c}, \tau-c ]$ with $c \sim \mathcal{U}\{1, \ldots, \tau-2\}$. 
In the Euclidean case, we use the DOPRI ODE solver~\cite{DormandPrince80:DOPRI} implemented in torchdyn~\cite{politorchdyn}. 
In the Riemannian case, we employ a Riemannian ODE solver based on the Euler method, as in~\cite{Chen24:RiemannianFM}.

\subsection{RFMP implementation}
Our RFMP implementation builds on the RFM framework from Chen and Lipman~\cite{Chen24:RiemannianFM}. 
Specifically, we parameterized the vector field $v_t(\bm{a}|\bm{o}; \bm{\theta})$ using a standard multilayer perceptron (MLP) with $64$ hidden units and $5$ layers for all experiments reported in the sequel.
We used the Swish activation function~\cite{Ramachandran17:Swish} with a learnable parameter.
The input to the MLP network is a vector formed as the concatenation of time and the observation vector. 
Under the aforementioned configuration, the resulting model has a total of $32$K learnable parameters. 
We optimized the network parameters $\bm{\theta}$ using Adam with a learning rate of $1e-4$ and an exponential moving averaging on the weights~\cite{Polyak92:MovingAverage} with a decay of $0.999$. For all experiments, we split the data as $80\%$ train, $10\%$ validation, and $10\%$ test. We trained the network for $200$ epochs and selected the best model based on its performance on the validation set.

Concerning the base distribution, our RFMP uses a Gaussian distribution $p_0 = \mathcal{N}(\bm{0}, \sigma\bm{I})$, in the Euclidean case $\manifold=\mathbb{R}^2$, where we set $\sigma=1$ for the experiments reported next. 
In the Riemannian setting, i.e., $\manifold=\sphere{2}$, we define the base distribution as a wrapped Gaussian distribution~\cite{Mardia99:DirectionalStats, Galaz-Garcia22:WrappedHomogenous} centered at the manifold origin $\bm{e}=(0, \ldots, 0, 1)^\trsp\in\mathcal{S}^d$, i.e., $p_0 = \mathcal{N}_{\sphere{2}}\big(\bm{e}, \sigma\bm{I}\big)$ with $\sigma=0.5$ for our experiments.

\section{EXPERIMENTS}
\label{sec:Experiments}
We evaluate RFMP on the LASA dataset~\cite{Lemme2015:LasaDataset} in the Euclidean space $\mathbb{R}^2$ and on the same dataset but projected on the sphere $\sphere{2}$. 
Using these datasets, we consider: (1) Trajectory-based policies, where observations are defined as current and past states along the trajectories; and (2) Visuomotor policies, where observations correspond to vector features extracted from grayscale images.
    
\subsection{Trajectory-based policies}
\label{subsec:trajectory-based-policies}
To learn the vector field $v_t$, we use a dataset  
$\{\{\{\bm{a}_{m,\tau}, \bm{o}_{m, \tau}\}_{c=1}^{\tau-2}\}_{\tau=2}^{T_m}\}_{m=1}^M$ 
of $M=7$ demonstrations containing $T_m=200$ timesteps each, where $\bm{a}_{m,\tau}=[\bm{a}_{m,\tau}, \ldots, \bm{a}_{m,\tau+T_a} ]$ and $\bm{o}_{m,\tau}=[\bm{o}_{m,\tau-1}, \bm{o}_{m,c}, \tau-c]$ are the action and observation vectors of the $\tau$-th step of the $m$-th demonstration. 
All actions and observations are normalized and projected onto the manifold $\manifold$ of interest. 
In this experiment, both actions $\bm{a}_\tau\in\manifold$ and observations $\bm{o}_\tau\in\manifold$ are represented in the position space of the manifold $\manifold$. 
The variance of the base distribution $p_0$ is selected such that the distribution roughly spans half of the sphere containing the data. 
We use a prediction horizon $T_a=8$ and an execution horizon $T_e = T_a/2$.

Figure~\ref{Fig:RMFPsamples} shows the learned RFMP flows conditioned on observations from the training dataset on the manifolds $\euclideanspace^2$ and $\sphere{2}$, for the letters $\mathsf{S}$ and $\mathsf{W}$. 
We observe that the distributions reconstructed by RFMP closely match the original demonstrations. 
Figure~\ref{subFig:RFMP-trajectories} displays the trajectories reconstructed by sequentially executing the actions inferred by RFMP. The obtained trajectories closely follow the demonstrations when initialized with the same initial observations. 
Notably, when tested on initial conditions that are randomly-sampled on the neighborhood of the demonstrations support, RFMP generates trajectories that closely follow the demonstrations pattern.
This kind of generalization is desired, for example, when the task demands to reproduce trajectories that closely resemble the demonstration style.

\begin{figure*}
    \centering
    \captionsetup{type=figure}
	\begin{subfigure}[b]{.48\linewidth}
        \centering
		\includegraphics[trim={2cm 2cm 2cm 3cm}, clip, width=.32\textwidth]{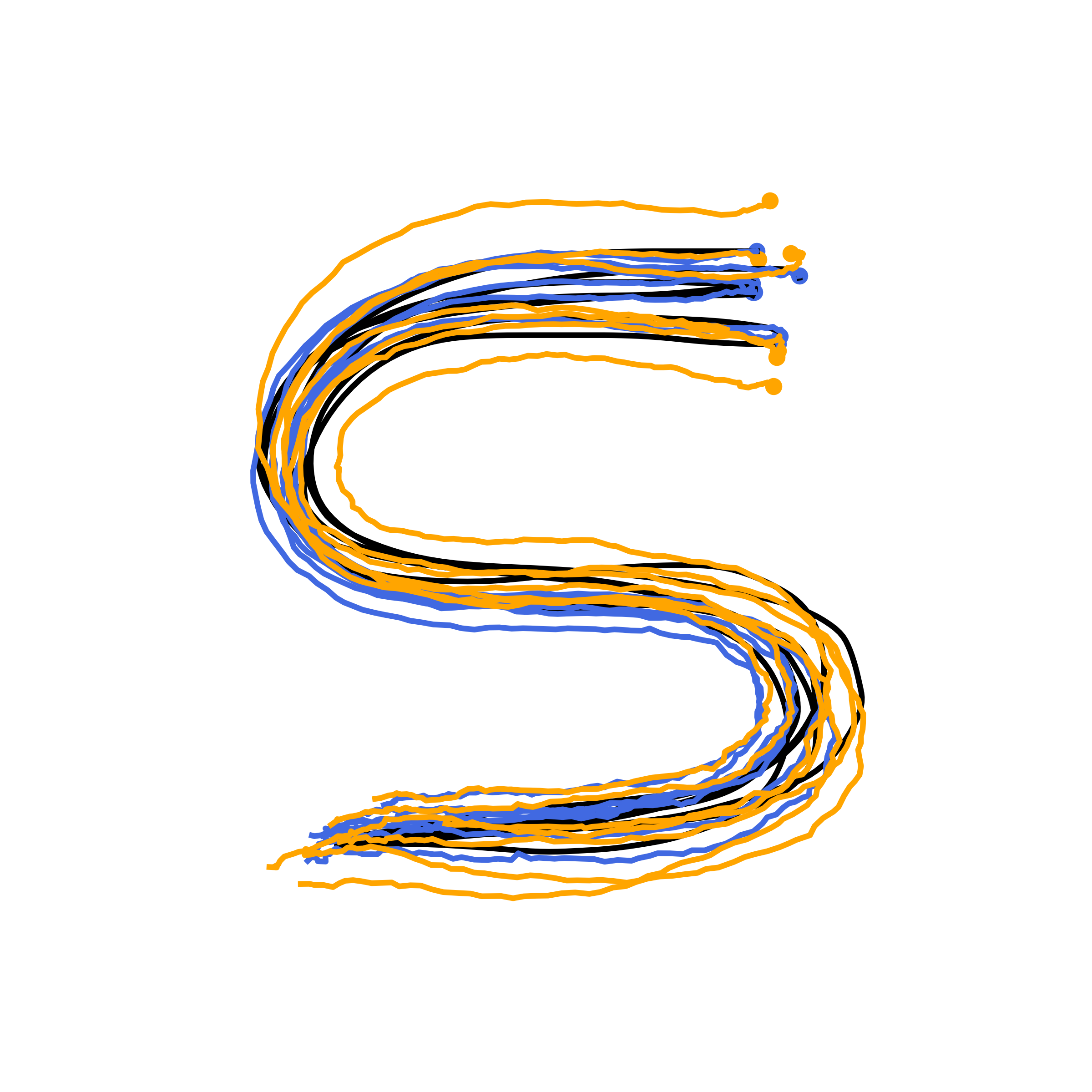}
    \includegraphics[trim={2cm 2cm 2cm 3cm}, clip, width=.32\textwidth]{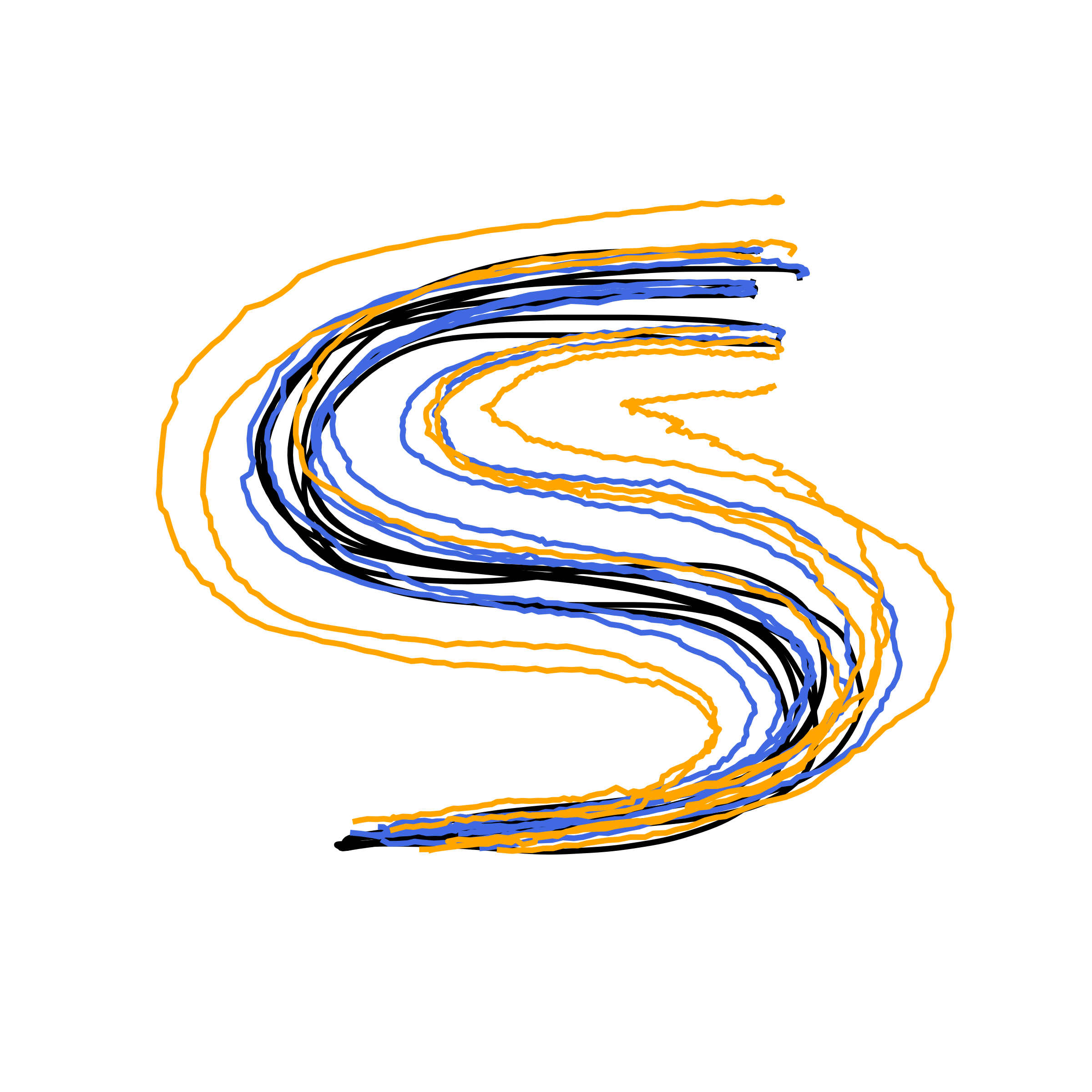}
        \includegraphics[trim={2cm 2cm 2cm 3cm}, clip, width=.32\textwidth]{Figures/trajectories/rfmp/lasa_euclidean_ref_cond_S_n8_r1_c1_w0_a4.png}
		\caption{RFMP on $\mathbb{R}^2$.} 
		\label{subFig:RFMPnpredR2}
	\end{subfigure}
    \begin{subfigure}[b]{.48\linewidth}
        \centering
		\includegraphics[trim={5cm 4cm 5cm 4cm}, clip,  width=.32\textwidth]{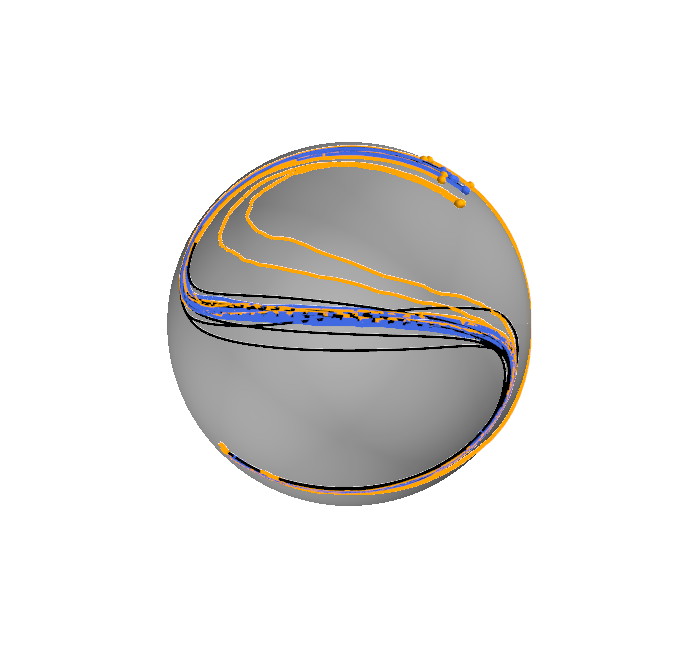}
        \includegraphics[trim={5cm 4cm 5cm 4cm}, clip,  width=.32\textwidth]{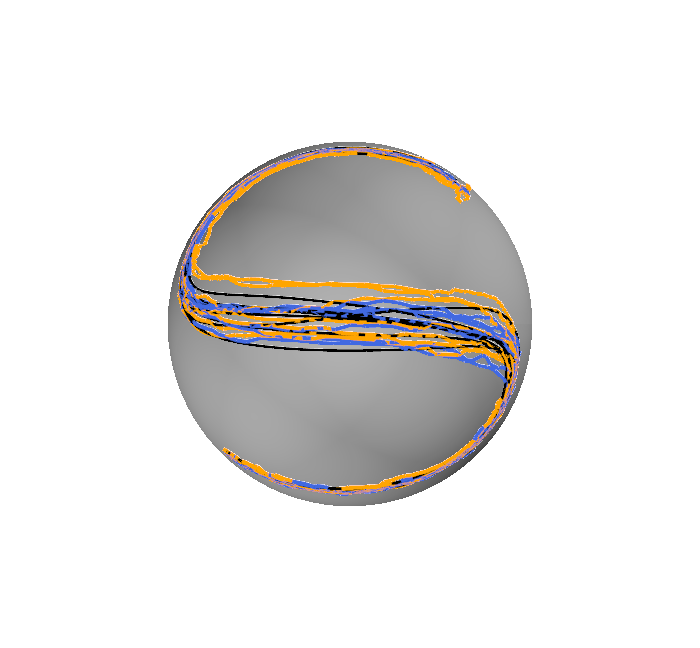}
        \includegraphics[trim={5cm 4cm 5cm 4cm}, clip,  width=.32\textwidth]{Figures/trajectories/rfmp/lasa_sphere_ref_cond_S_n8_r1_c1_w0_a4.png}
		\caption{RFMP on $\mathcal{S}^2$.} 
		\label{subFig:DPnpredR2}
	\end{subfigure}
    \begin{subfigure}[b]{.48\linewidth}
        \centering
		\includegraphics[trim={2cm 2cm 2cm 3cm}, clip, width=.32\textwidth]{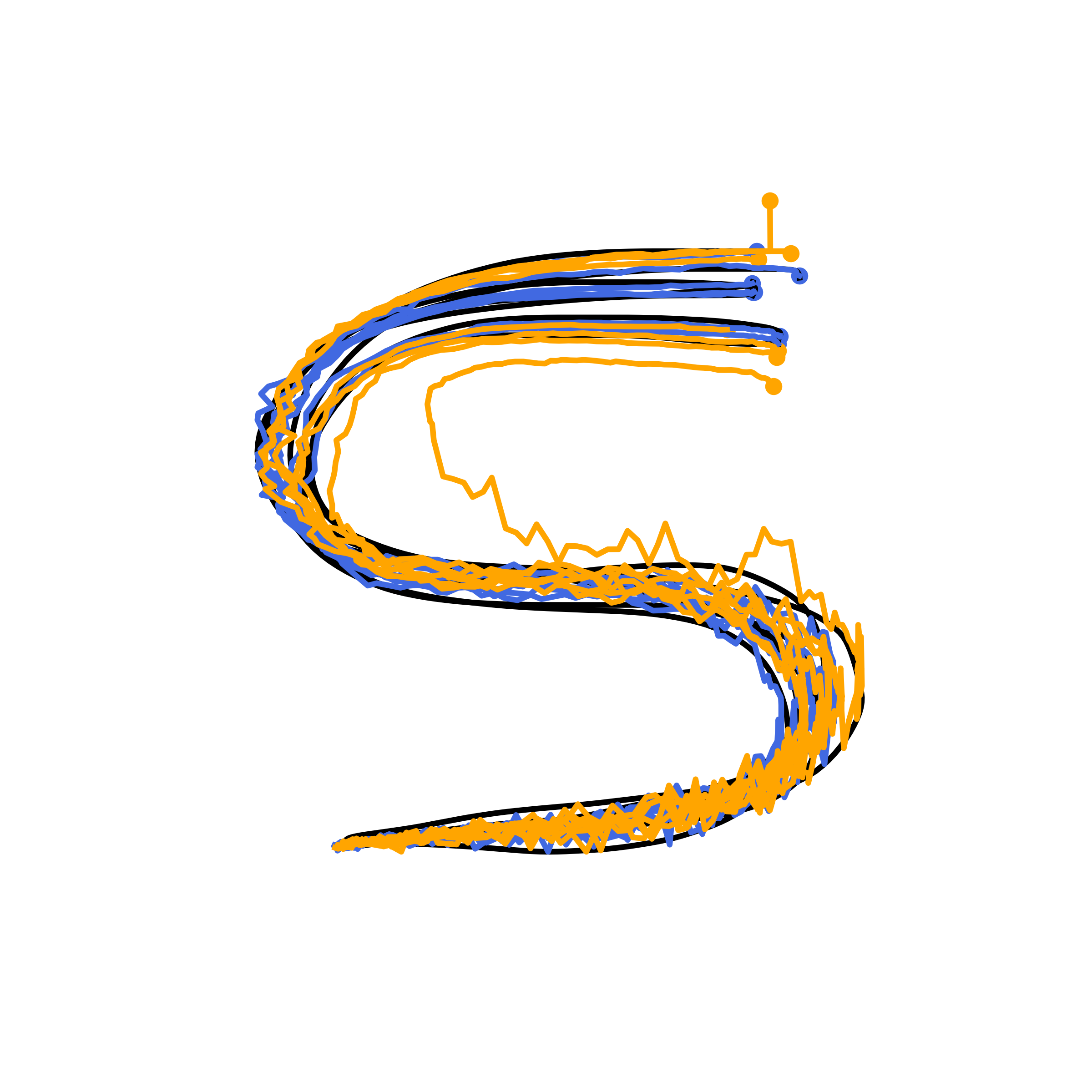}
    \includegraphics[trim={2cm 2cm 2cm 3cm}, clip, width=.32\textwidth]{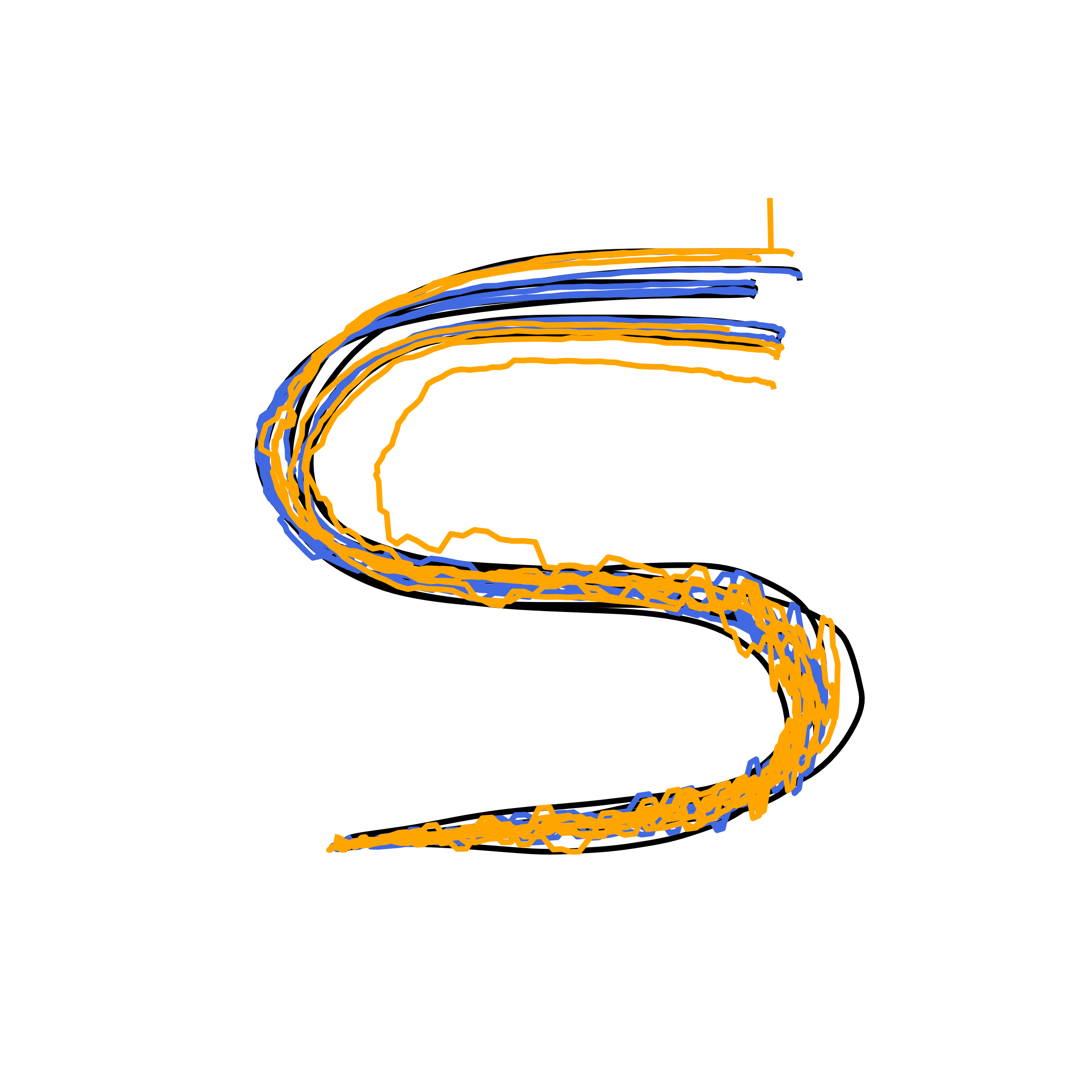}
        \includegraphics[trim={2cm 2cm 2cm 3cm}, clip, width=.32\textwidth]{Figures/trajectories/dp/lasa_euclidean_ref_cond_S_a4.png}
		\caption{DP on $\mathbb{R}^2$.} 
		\label{subFig:RFMPnpredS2}
	\end{subfigure}
    \begin{subfigure}[b]{.48\linewidth}
        \centering
		\includegraphics[trim={5cm 4cm 5cm 4cm}, clip,  width=.32\textwidth]{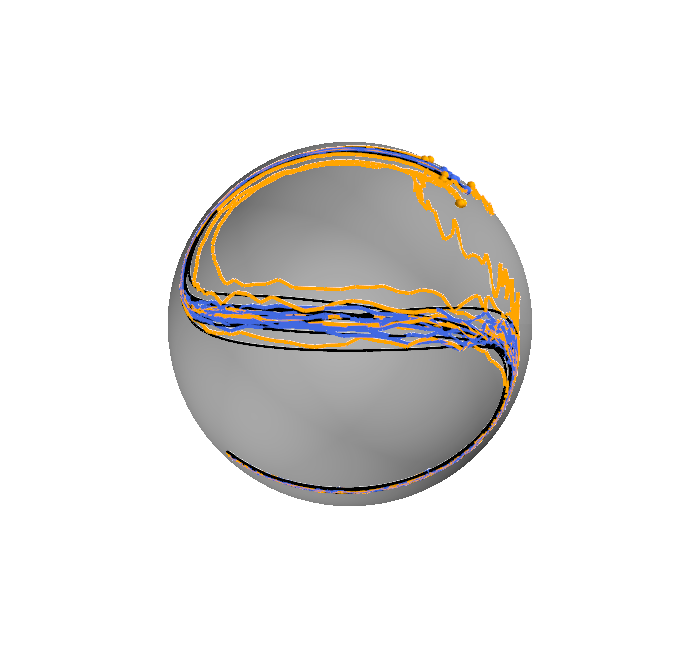}
        \includegraphics[trim={5cm 4cm 5cm 4cm}, clip,  width=.32\textwidth]{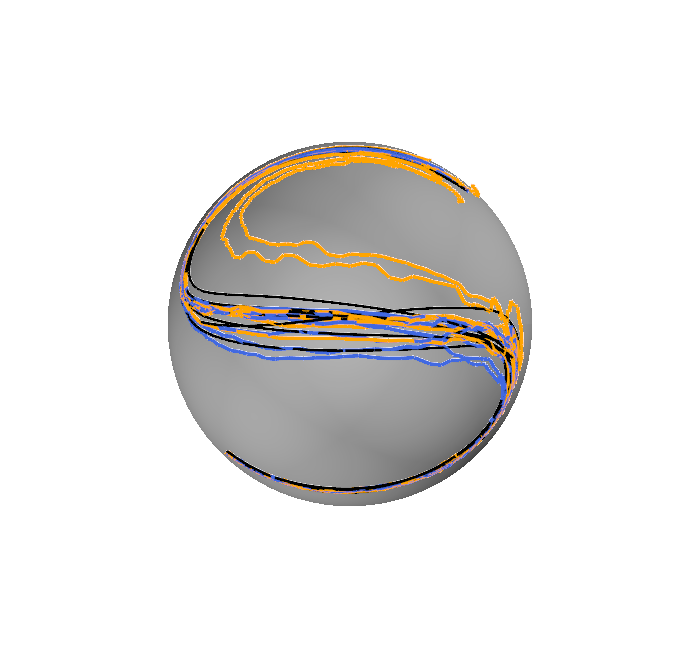}
        \includegraphics[trim={5cm 4cm 5cm 4cm}, clip,  width=.32\textwidth]{Figures/trajectories/dp/lasa_sphere_ref_cond_S_a4.png}
		\caption{DP on $\mathcal{S}^2$.} 
		\label{subFig:DPnpredS2}
	\end{subfigure}
	\caption{Demonstrations (\blackline) and learned trajectories on the LASA datasets $\mathsf{S}$ and $\mathsf{W}$ with different prediction horizons $T_a=\{2,4,8\}$ (from left to right). Reproductions start at the same initial observations as the demonstrations (\royalblueline), or from randomly-sampled observations in their neighborhood (\orangeline).}
	\label{Fig:RMFP-trajectories-npred}
    \vspace{-0.3cm}
\end{figure*}

We compare RFMP against diffusion policies (DP)~\cite{Cheng23:DiffPolicy}. 
To do so, we employ the CNN-based diffusion network with $256$M parameters provided by the authors. 
As in~\cite{Cheng23:DiffPolicy}, we used the iDDPM algorithm~\cite{Nichol21:ImprovedDenoisingDPM} with the same $100$ denoising diffusion iterations for both training and inference. 
Moreover, we used the same prediction and execution horizons as for RFMP.
Figure~\ref{subFig:DP-trajectories} shows the trajectories obtained by sequentially executing the actions inferred by DP. 
Similarly to RFMP, the trajectories closely match the demonstrations when initialized with the same initial observations (blue curves). 
Interestingly, unlike RFMP, the trajectories starting  at randomly-sampled initial conditions close to the demonstrations (orange curves), tend to rejoin the demonstration data support, resulting in less variance across reproductions. 
This behavior might be partly explained by a high memorization of the training data~\cite{Yoon23:DiffusionMemorization}, although this requires further investigation.

Importantly, the trajectories obtained by DP tend to be more jerky than those obtained with RFMP. 
We hypothesize that such jerky trajectories are a result of the inherent stochasticity of diffusion models during inference. 
These observations are supported by quantitative measures. Table~\ref{tab:QuantitativeMeasures-trajectories} shows the dynamic time warping distance (DTWD) as a measure of reproduction accuracy for trajectories initialized with the same initial observations as the demonstrations, and the jerkiness as a measure of the trajectories smoothness~\cite{Balasubramanian15:Smoothness}. 
We observe that DP produces trajectories that display a similar or lower DTWD than RFMP. 
This can be explained by the tendency of DP to generate trajectories within the demonstrations support, while RFMP displays an increased variance across reproductions. 
As observed qualitatively, RFMP produces arguably smoother trajectories than DP, as indicated by the lower jerkiness values reported in Table~\ref{tab:QuantitativeMeasures-trajectories}.

Note that diffusion policies are not adapted to handle data on Riemannian manifolds, and thus do not provide any guarantees that the resulting trajectories lie on the manifold of interest. 
This can be observed, e.g., for the $\mathsf{S}$ dataset on $\sphere{2}$, where some trajectories enter the sphere in the middle part of the  $\mathsf{S}$ trajectories. 
This means that a post-processing step would be required to ensure that the trajectories lie on the manifold of interest. 
Although possible, such post-processing steps are known to produce highly-inaccurate predictions as they disregard the intrinsic geometry of the data, as discussed in~\cite{Jaquier24:Fallacy}.
A more technically-sound solution would involve to adapt diffusion policies using Riemannian formulations of diffusion models~\cite{Huang22:RiemannianDM,Lou23:ScalingRiemannianDM}.

We also tested the capabilities of both RFMP and DP to learn multimodal policies on Riemannian manifolds.
The rightmost plots in Fig.~\ref{Fig:RMFP-trajectories} show the resulting trajectories for initial conditions matching those of the demonstrations dataset (blue curves) and for initial points drawn from a region close to the demonstrations (orange curves). 
Although both RFMP and DP are able to learn the multimodal pattern, their generalization behavior is different when tested on initial conditions that are different from the training dataset. 
We can again observe that the DP trajectories tend to move back to the data support. 
Interestingly, in the multimodal case, RFMP outperforms DP in terms of both DTWD and smoothness (see Table~\ref{tab:QuantitativeMeasures-trajectories}).

Next, we ablate the prediction horizon $T_a$ for both RFMP and DP. 
Figure~\ref{Fig:RMFP-trajectories-npred} shows the trajectories obtained for both policy types with $T_a=\{2,4,8\}$ and $T_e = T_a/2$. 
The corresponding quantitative measures (DTWD and jerkiness) are given in Table~\ref{tab:QuantitativeMeasures-trajectories-npred}. Interestingly, the RFMP trajectories remain smooth despite the reduction of the prediction horizon, even for $T_a=2$. 
In contrast, DP exhibits jerkier behaviors with shorter prediction horizons. 
This trend is especially pronounced for trajectories starting from initial observations that are different from the training dataset (see orange curves in Fig.~\ref{subFig:DPnpredR2}-\emph{left} and Fig.~\ref{subFig:DPnpredS2}-\emph{left}). 
We hypothesize that the reduction of the prediction horizon has a stronger impact on the inference process of DP than on RFMP due to the inherent stochasticity of diffusion models. 

Finally, we compare the inference time of RFMP and DP in Table~\ref{tab:InferenceTimes}. 
All computations were performed on a laptop with $2.60$GHz $\times12$ CPU, a Nvidia Quatro T200 GPU, and $31$ GB RAM.
In the Euclidean case, we observe a reduction of $\sim30\%$ ($\sim 350$ms) for the inference time of RFMP compared to DP. 
This is due to the fact that RFMP employs ODE solvers, which are generally much more efficient that the SDE solvers required for diffusion models. 
This result is an inherent strength of flow matching compared to diffusion models, as thoroughly analyzed in the flow matching literature~\cite{Lipman23:FlowMatching,Tong23:CFMoptimaltransport}. 
The inference time of RFMP on the sphere $\sphere{2}$ exceeds that of DP. 
However, this is due to the fact that RFMP intrinsically handles data on Riemannian manifolds and therefore uses ODE solvers that are specific to this type of spaces. 
Instead, DP disregards the geometry of the data and thus employs Euclidean SDE solvers. 
A fair comparison of the inference times of RFMP and DP on the sphere would involve the adaptation of diffusion policies to data on Riemannian manifolds and leveraging Riemannian SDE solvers for inference.

\begin{table*}[t]
    \setlength{\tabcolsep}{2pt}
    \renewcommand{\arraystretch}{1.05}
    \centering
    \begin{tabular}{c|cccc|cccc|}
        \multicolumn{1}{c|}{} & \multicolumn{4}{c|}{DTWD} &\multicolumn{4}{c|}{Jerkiness}\\
        Dataset & $\mathsf{S}$, $\euclideanspace^2$ & $\mathsf{S}$, $\sphere{2}$ &$\mathsf{W}$, $\sphere{2}$ & multi-$\mathsf{L}$, $\sphere{2}$ & $\mathsf{S}$, $\euclideanspace^2$ & $\mathsf{S}$, $\sphere{2}$ &$\mathsf{W}$, $\sphere{2}$ & multi-$\mathsf{L}$, $\sphere{2}$ \\
         \hline
         RFMP & $1.87\pm0.94$ & $0.95\pm0.32$ & $1.64\pm0.84$ & $\bm{6.14\pm6.56}$ & $\bm{2120\pm273}$ & $\bm{4077\pm900}$ & $4198\pm560$ & $\bm{2161\pm640}$\\  
		DP &$\bm{0.98\pm0.22}$ & $\bm{0.80\pm0.21}$ & $\bm{0.90\pm0.35}$ & $7.06\pm7.73$ &$8172\pm747$ & $7612\pm543$ & $\bm{2944\pm1399}$ & $2201\pm744$ \\
	    \hline
 \end{tabular}
    \caption{Average dynamic time warping distance (DTWD) and jerkiness (a.k.a smoothness) for trajectory-based RFMP and DP. DTWD is computed between the demonstrations and the reproductions initialized as the demonstrations, while the smoothness is averaged over all the reproductions displayed in Fig.~\ref{Fig:RMFP-trajectories}.}
    \label{tab:QuantitativeMeasures-trajectories}
\end{table*}

\begin{table*}[t]
    \setlength{\tabcolsep}{2pt}
    \renewcommand{\arraystretch}{1.05}
    \centering
    \resizebox{\textwidth}{!}{
    \begin{tabular}{c|ccc|ccc|ccc|ccc|}
        \multicolumn{1}{c|}{} & \multicolumn{3}{c|}{DTWD, $\euclideanspace^2$} &\multicolumn{3}{c|}{Jerkiness, $\euclideanspace^2$}  & \multicolumn{3}{c|}{DTWD, $\sphere{2}$} &\multicolumn{3}{c|}{Jerkiness, $\sphere{2}$}\\
        $T_a$ & $2$ & $4$ & $8$ & $2$ & $4$ & $8$ & $2$ & $4$ & $8$ & $2$ & $4$ & $8$ \\
         \hline
         RFMP & $1.13\pm0.29$ & $2.31\pm 1.25$ & $1.87\pm0.94$ & $\bm{3454\pm 276}$& $\bm{3729\pm 475}$& $\bm{2120\pm273}$ & $\bm{0.52\pm 0.18}$ & $0.90\pm 0.34 $&$0.95\pm0.32$ & $\bm{2845\pm 335}$& $6169\pm 556$ & $\bm{4077\pm900}$ \\  
		DP & $\bm{0.70\pm 0.07}$& $\bm{0.76\pm 0.25}$& $\bm{0.98\pm0.22}$ & $11905\pm 1962$& $7289\pm 939$ & $8172\pm747$ & $0.60\pm 0.16$ & $\bm{0.70\pm0.33}$ &$\bm{0.80\pm0.21}$ & $6586\pm3140$ & $\bm{4239\pm 1306}$& $7612\pm543$ \\
	    \hline
 \end{tabular}
 }
    \caption{Average dynamic time warping distance (DTWD) and jerkiness (a.k.a smoothness) for RFMP and DP with different prediction horizons $T_a=\{2,4,8\}$. DTWD is computed between the demonstrations and the reproductions initialized as the demonstrations, while the smoothness is averaged over all the reproductions displayed in Fig.~\ref{Fig:RMFP-trajectories-npred}.}
    \label{tab:QuantitativeMeasures-trajectories-npred}
\end{table*}

\begin{table}[t]
    \setlength{\tabcolsep}{2pt}
    \renewcommand{\arraystretch}{1.05}
    \centering
    \begin{tabular}{c|cc|cc|cc|cc}
        \multicolumn{1}{c|}{} & \multicolumn{2}{c|}{Trajectory-based} & \multicolumn{2}{c|}{Visuomotor} \\
        Dataset & $\mathsf{S}$, $\euclideanspace^2$ & $\mathsf{S}$, $\sphere{2}$ &  $\mathsf{S}$, $\euclideanspace^2$ & $\mathsf{S}$, $\sphere{2}$ \\
         \hline
         RFMP & $\bm{803 \pm 55}$ & $1539 \pm 23$ & $\bm{1355 \pm110}$&$\bm{2351 \pm 88}$ \\  
		  DP &$1142 \pm 17$ & $\bm{1147 \pm 26}$ & $2462 \pm141$ & $2662\pm541$\\
	    \hline
 \end{tabular}
    \caption{Inference times (in milliseconds) per prediction step for RFMP and DP. These are averaged across the $50$ prediction steps with $T_a=8$, for both the $14$ reproductions displayed in Fig.~\ref{Fig:RMFP-trajectories} for trajectory-based policies, and the $7$ reproductions displayed in Fig.~\ref{Fig:RMFP-visuomotor-trajectories} for visuomotor policies.}
    \label{tab:InferenceTimes}
    \vspace{-0.3cm}
\end{table}

\subsection{Towards visuomotor policies}

\begin{figure}[t]
    \centering
    \captionsetup{type=figure}
	\begin{subfigure}[b]{.47\linewidth}
		\includegraphics[trim={3.0cm 1.5cm 3.0cm 1.0cm}, clip,width=\textwidth]{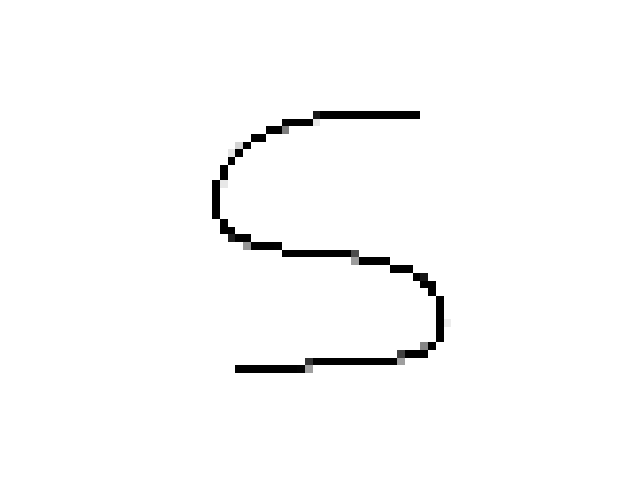}
		\caption{$\mathbb{R}^2$} 
		\label{subFig:EuclideanSamples}
	\end{subfigure}
	\begin{subfigure}[b]{.47\linewidth}
		\includegraphics[trim={4.0cm 1.5cm 0.5cm 1.0cm}, clip, width=\textwidth]{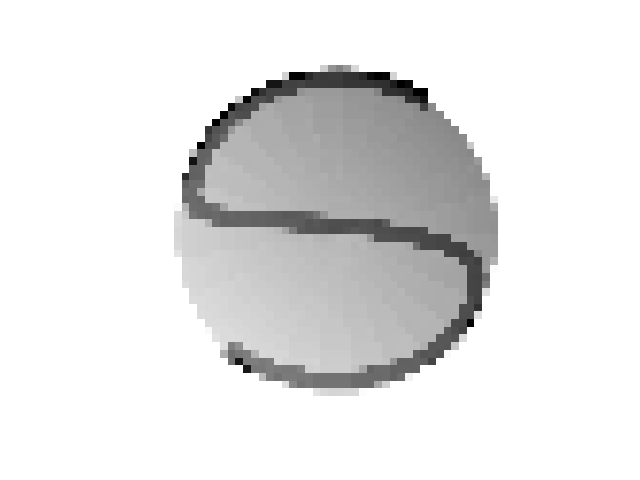}
		\caption{$\mathcal{S}^2$} 
		\label{subFig:SphereSamples}
	\end{subfigure}
	\caption{Examples of visual observations at the end of a demonstration of the LASA dataset $\mathsf{S}$.}
	\label{Fig:RMFPvision}
    \vspace{-0.3cm}
\end{figure}

In this section, we study the case where the RFMP vector field is conditioned on visual observations, thus resembling a visuomotor diffusion policy. 
Similarly to Section~\ref{subsec:trajectory-based-policies}, we use a dataset defined as
$\{\{\{\bm{a}_{m,\tau}, \bm{o}_{m, \tau}\}_{c=1}^{\tau-2}\}_{\tau=2}^{T_m}\}_{m=1}^M$ 
of $M=7$ demonstrations containing $T_m=200$ timesteps each, where $\bm{a}_{m,\tau}=[\bm{a}_{m,\tau}, \ldots, \bm{a}_{m,\tau+T_a} ]$ and $\bm{o}_{m,\tau} = [\bm{o}_{m,\tau-1}, \bm{o}_{m,c}, \tau-c]$ are the action and observation vectors of the $\tau$-th step of the $m$-th demonstration. 
Again, all action and observation vectors are normalized and projected onto the manifold $\manifold$ of interest. 
However, in this case, the observations $\bm{o}_\tau\in\manifold$ are given by the latent encodings (a.k.a feature vectors) of $48\times48$ raw grayscale images depicting the temporal progress of the task. 
Examples of such images are shown in Fig.~\ref{Fig:RMFPvision}.
Specifically, our vision perception backbone, which maps raw grayscale images to observation vectors, is exactly the same used in DP~\cite{Cheng23:DiffPolicy}. 
Namely, we used a standard ResNet-$18$ in which we replaced: (1) the global average pooling with a spatial softmax pooling, and (2) BatchNorm with GroupNorm. 
The former modification maintains spatial information~\cite{Mandlekar22:LearningFromOfflineDemos}, while the latter stabilizes the training~\cite{Wu18:GroupNormalization}.

\begin{figure*}[t]
    \centering
    \captionsetup{type=figure}
	\begin{subfigure}[b]{\linewidth}
        \centering
		\includegraphics[trim={1cm 2cm 1cm 3cm}, clip,width=.22\textwidth]{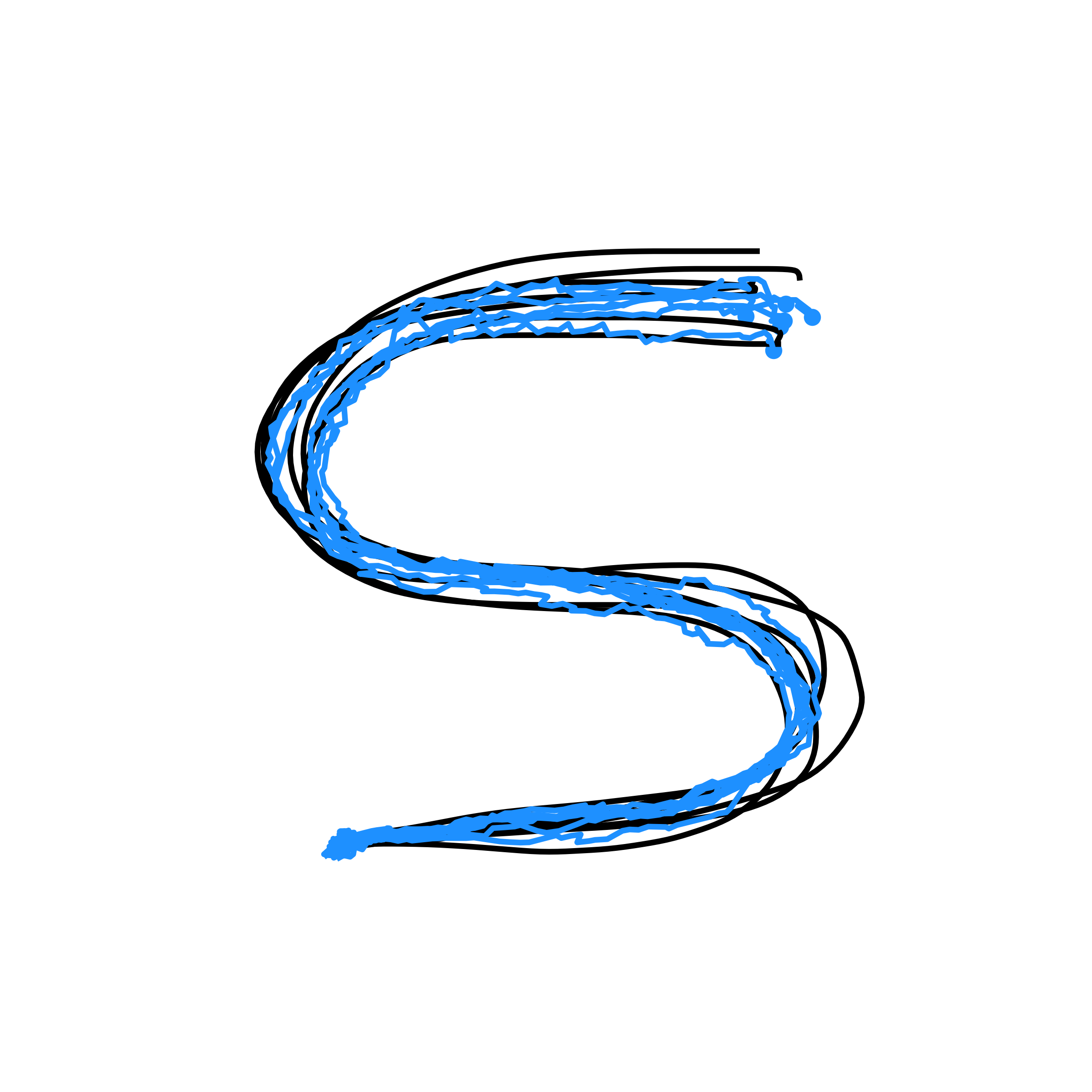}
        \includegraphics[trim={1cm 2cm 1cm 3cm}, clip,width=.22\textwidth]{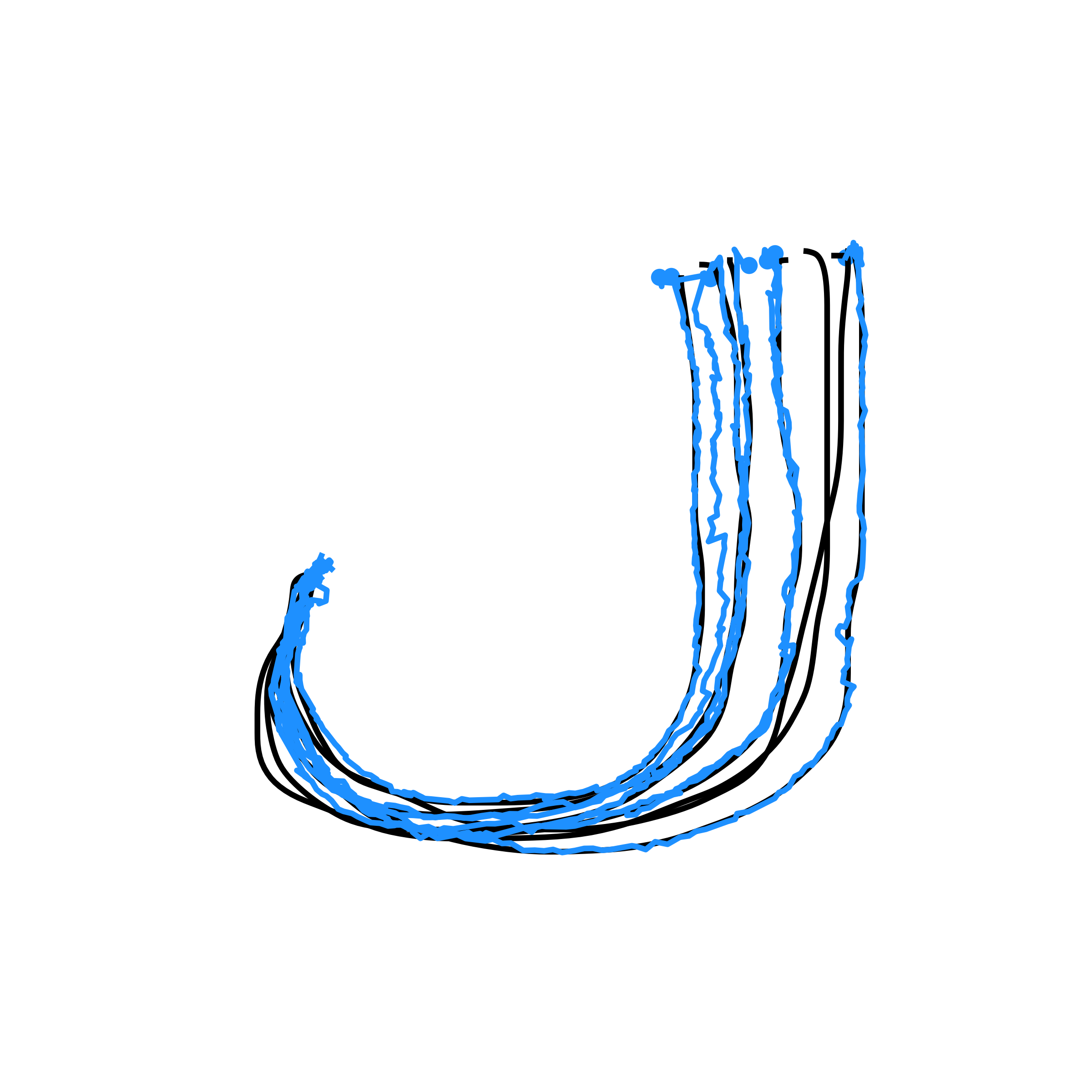}
        \includegraphics[trim={4cm 4cm 4cm 4cm}, clip, width=.22\textwidth]{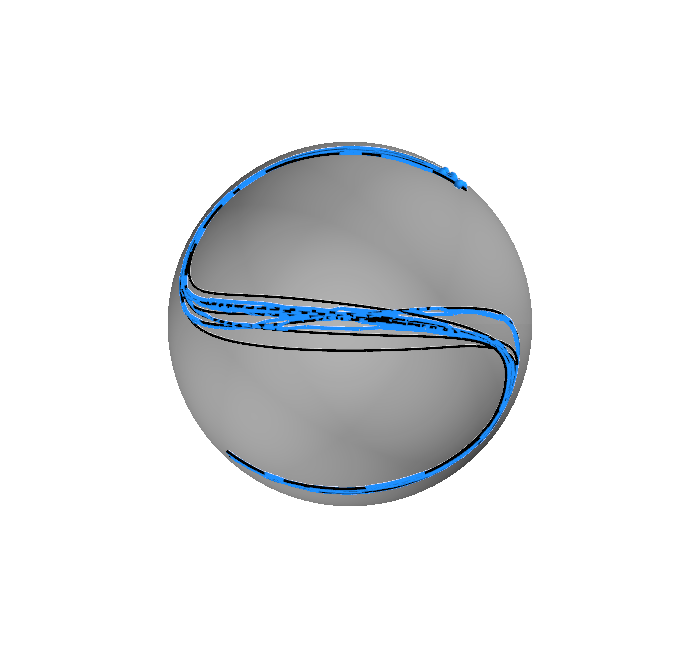}
        \includegraphics[trim={4cm 4cm 4cm 4cm}, clip, width=.22\textwidth]{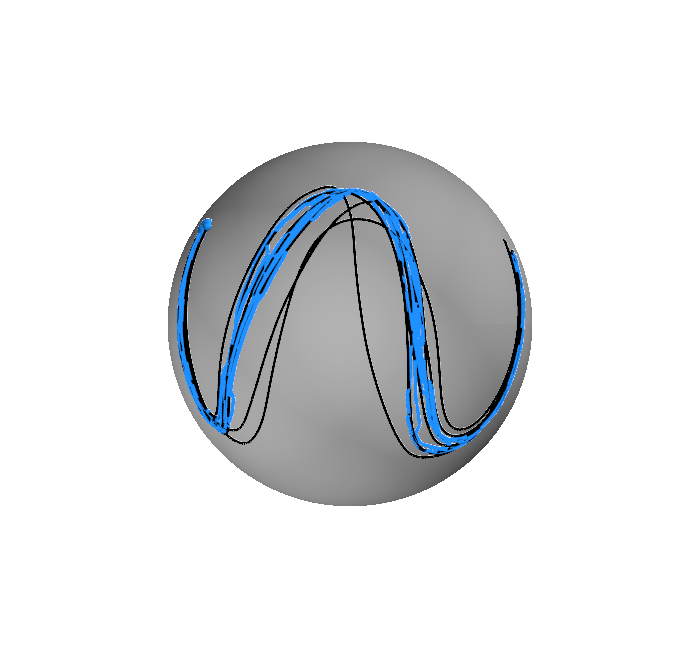}
		\caption{RFMP trajectories.} 
		\label{subFig:RFMPvisuomotor-traj}
	\end{subfigure}
    \begin{subfigure}[b]{\linewidth}
        \centering
		\includegraphics[trim={1cm 2cm 1cm 3cm}, clip,width=.22\textwidth]{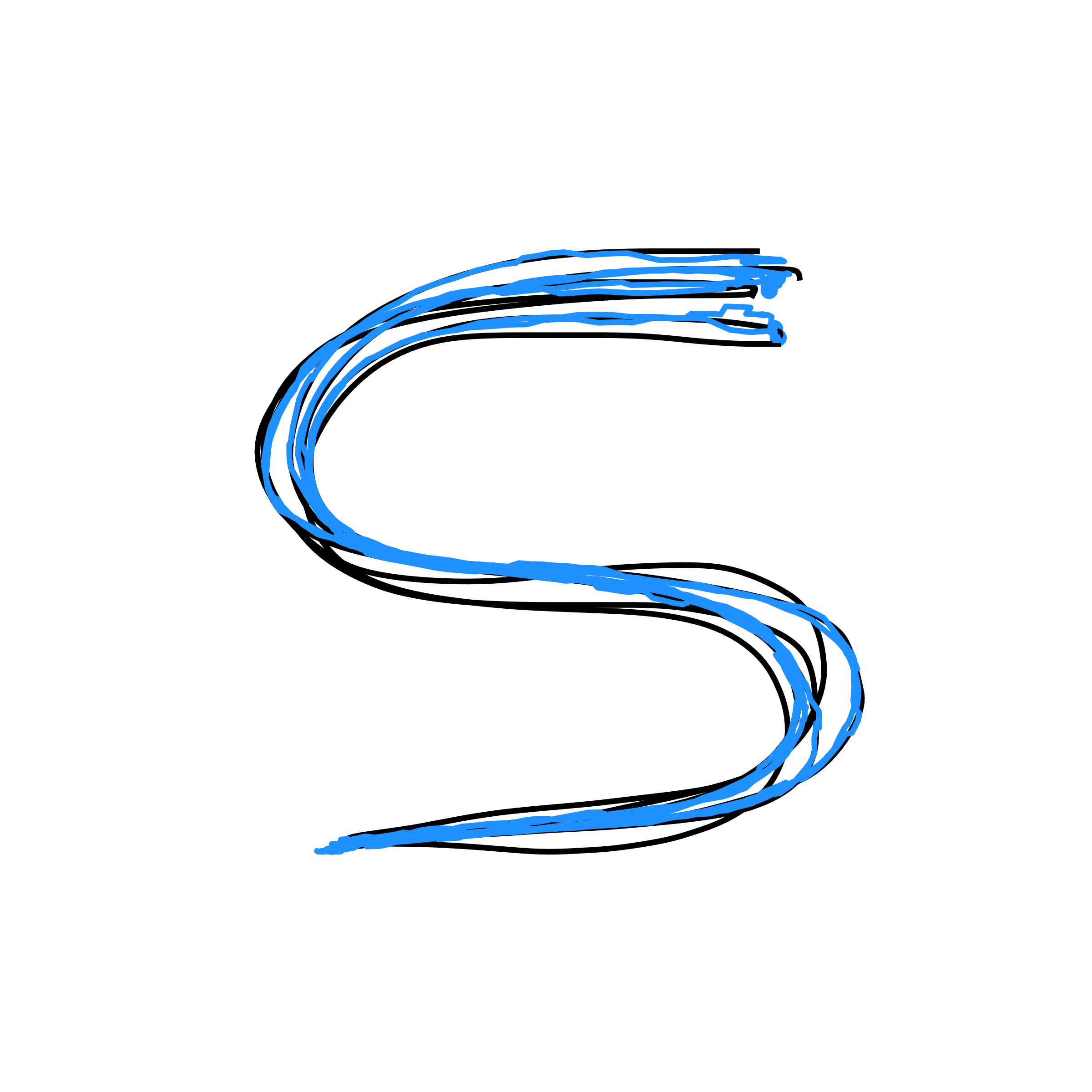}
        \includegraphics[trim={1cm 2cm 1cm 3cm}, clip,width=.22\textwidth]{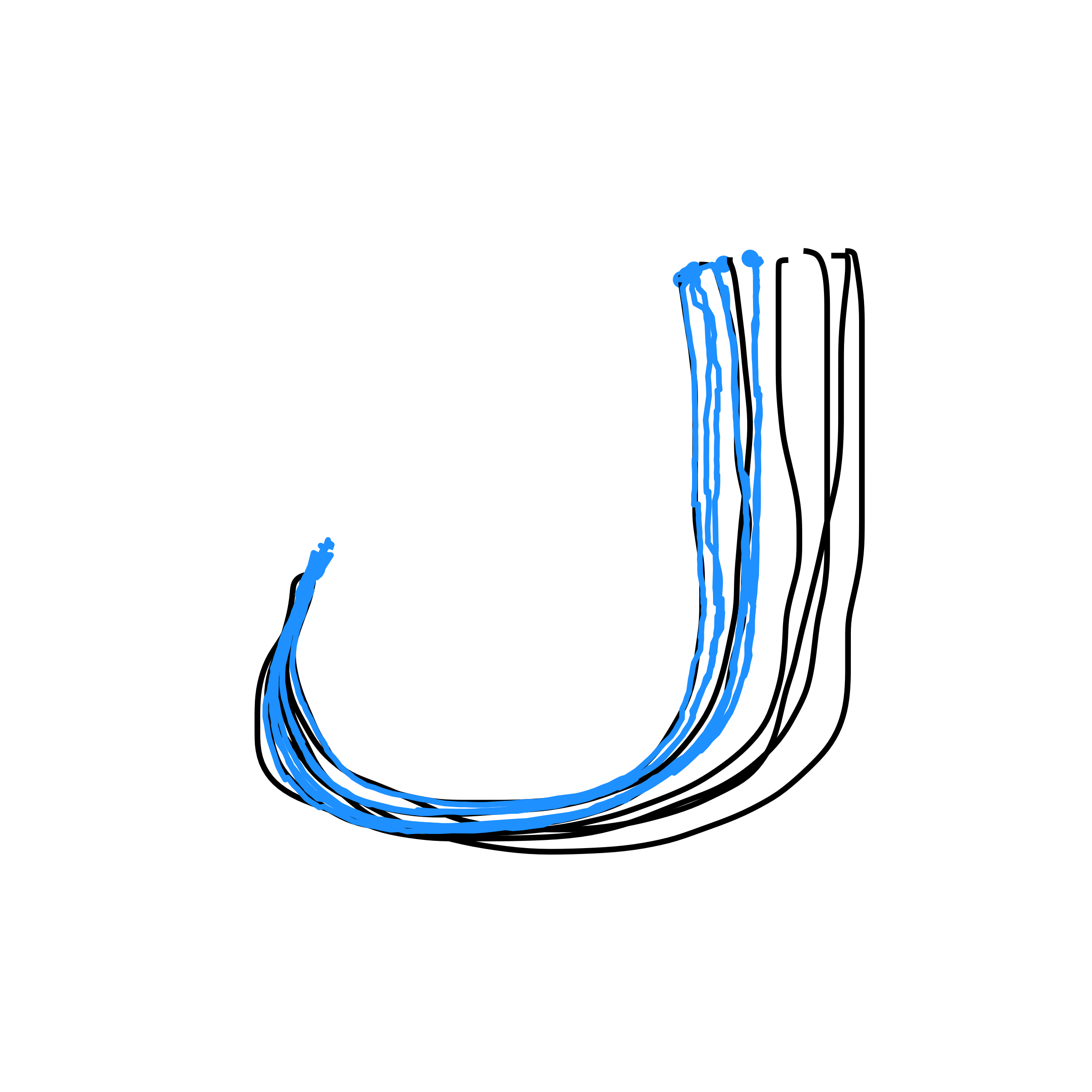}
        \includegraphics[trim={4cm 4cm 4cm 4cm}, clip, width=.22\textwidth]{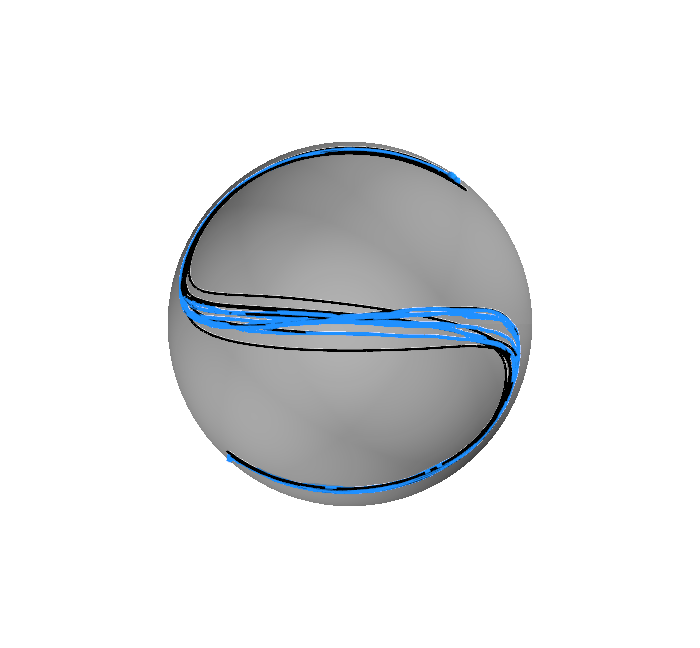}
        \includegraphics[trim={4cm 4cm 4cm 4cm}, clip, width=.22\textwidth]{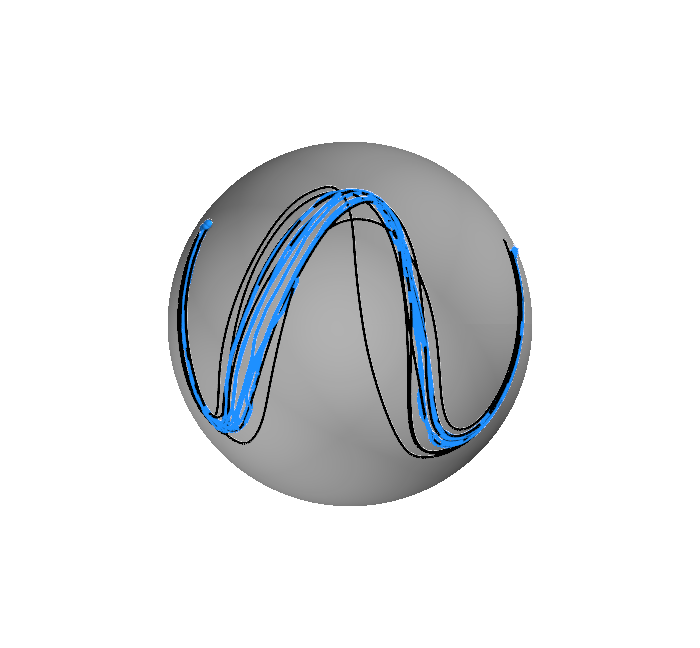}
		\caption{DP trajectories.} 
		\label{subFig:DPvisuomotor-traj}
	\end{subfigure}
	\caption{Demonstrations (\blackline) and trajectories reproduced by the visuomotor RFMP and DP (\dodgerblueline) on the LASA datasets $\mathsf{S}$ and $\mathsf{J}$ in $\euclideanspace^2$ (\emph{left}) and on the LASA datasets $\mathsf{S}$ and $\mathsf{W}$ projected on $\sphere{2}$ (\emph{right}).}
	\label{Fig:RMFP-visuomotor-trajectories}
\end{figure*}

\begin{table*}[t]
    \setlength{\tabcolsep}{2pt}
    \renewcommand{\arraystretch}{1.05}
    \centering
    \begin{tabular}{c|cccc|cccc|}
        \multicolumn{1}{c|}{} & \multicolumn{4}{c|}{DTWD} &\multicolumn{4}{c|}{Jerkiness}\\
        Dataset & $\mathsf{S}$, $\euclideanspace^2$ & $\mathsf{J}$, $\euclideanspace^2$ &$\mathsf{S}$, $\sphere{2}$ & $\mathsf{W}$, $\sphere{2}$ & $\mathsf{S}$, $\euclideanspace^2$ & $\mathsf{J}$, $\euclideanspace^2$ &$\mathsf{S}$, $\sphere{2}$ & $\mathsf{W}$, $\sphere{2}$ \\
         \hline
         RFMP & $\bm{1.22\pm0.44}$ & $\bm{1.82\pm0.93}$ & $0.76\pm0.27$ & $\bm{0.84\pm0.48}$ & $10543\pm612$ & $7655\pm537$ & $\bm{3590\pm353}$ & $\bm{4455\pm306}$\\  
		DP &$1.29\pm0.49$ & $2.35\pm1.66$ & $\bm{0.67\pm0.24}$ & $0.93\pm0.48$ &$\bm{6198\pm755}$ & $\bm{5588\pm801}$ & $5903\pm170$ & $5042\pm136$ \\
	    \hline
 \end{tabular}
    \caption{Average dynamic time warping distance (DTWD) and jerkiness (a.k.a smoothness) for visuomotor RFMP and DP. DTWD is computed between the demonstrations and the reproductions displayed in Fig.~\ref{Fig:RMFP-visuomotor-trajectories} and the smoothness is averaged over the same reproductions.}
    \label{tab:QuantitativeMeasures-visuomotor}
\end{table*}

We trained the visual encoder end-to-end with our RFMP, for which we used the same base distributions $p_0$ and prediction horizon $T_a=8$ as in Section~\ref{subsec:trajectory-based-policies}. 
In this visuomotor RFMP, we empirically observed that shortening the observations horizon used to sample the context observation increases temporal consistency and improves the smoothness on the predicted actions. 
Therefore, in the following experiments, we sample $c \sim \mathcal{U}\{\tau-w, \ldots, \tau-2\}$ with $w=50$. 
Figure~\ref{subFig:RFMPvisuomotor-traj} shows the demonstrations and the reproduced trajectories of the learned visuomotor RFMP. 
For both demonstrations and reproductions, the initial observations correspond to blank images for policies trained in $\euclideanspace^2$, and an empty grayscale sphere for policies in $\sphere{2}$. 
Similarly to the trajectory-based policies, the visuomotor RFMP successfully reproduces trajectories that match the demonstrations pattern in both the Euclidean and the Riemannian settings. 

We compare visuomotor RFMP against visuomotor DP. As in~\cite{Cheng23:DiffPolicy}, and similarly to the visuomotor RFMP, we train the vision perception backbone (modified ResNet-$18$) end-to-end with the CNN-based diffusion network described in Section~\ref{subsec:trajectory-based-policies}. Figure~\ref{subFig:DPvisuomotor-traj} shows the trajectories obtained by sequentially executing the actions inferred by the visuomotor DP. Similarly to RFMP, the trajectories closely match the demonstrations. Interestingly, the visuomotor RFMP competitively performs when compared to the visuomotor DP in terms of the DTWD metric, as shown in Table~\ref{tab:QuantitativeMeasures-visuomotor}, despite having a simpler architecture parametrizing the RFMP vector field.
Moreover, as observed in Section~\ref{subsec:trajectory-based-policies} for the trajectory-based case, visuomotor RFMP leads to smooth trajectories, especially for policies on $\sphere{2}$ as indicated by the low jerkiness values in Table~\ref{tab:QuantitativeMeasures-visuomotor}. Let us emphasize once more that RFMP ensures that the predicted actions lie on the manifold of interest, as opposed to DP which does not provide such guarantees.

Finally, we compare the inference time of visuomotor RFMP and DP in Table~\ref{tab:InferenceTimes}. We observe a reduction of $\sim 45\%$ ($\sim900$ms) for the inference time of RFMP compared to DP. Interestingly, this reduction is greater for visuomotor policies compared to the trajectory-based case. Moreover, the inference time of RFMP on the sphere $\sphere{2}$ is similar to that of DP, despite that RFMP uses Riemannian-specific ODE solvers which are computationally more expensive than Euclidean ODE solvers. 
The reported findings indicate comparable performance between RFMP and DP in terms of task completion. However, RFMP exhibits a clear advantage in generating smoother action predictions, and this advantage remains consistent regardless of the prediction horizon. Furthermore, RFMP boasts significantly faster inference times compared to DP.
These attributes make RFMP a compelling choice for real-time applications in various robotic domains.

\section{CONCLUSION}
We introduced Riemannian Flow Matching Policies (RFMP), a novel learning framework that leverages the simplicity and fast inference of flow matching models to model visuomotor robot policies on Riemannian manifolds. 
We evaluated RFMP on both trajectory-based and vision-based settings using the LASA dataset. 
Our results demonstrated that RFMP successfully learns policies that reproduce the demonstration patterns even for initial conditions outside the training data.
Compared to Diffusion Policies (DP), RFMP generates smoother predicted trajectories with significantly lower inference times. 
Interestingly, RFMP exhibited less performance degradation with decreasing action prediction horizons.
Notably, RFMP achieved this competitive performance using a simple MLP architecture for its vector field, in contrast to the more powerful CNN architecture employed by DP in score matching.
Our proof-of-concept experiments showed the potential of RFMP for learning complex visuomotor policies in real-world robotic applications. 
Future work will evaluate the performance of RFMP in real-world robotics applications. Moreover, we will explore more powerful representations for the RFMP vector field and more informative prior models. 

\bibliographystyle{IEEEtran}

\bibliography{References} 

\begin{thebibliography}{10}
\providecommand{\url}[1]{#1}
\csname url@samestyle\endcsname
\providecommand{\newblock}{\relax}
\providecommand{\bibinfo}[2]{#2}
\providecommand{\BIBentrySTDinterwordspacing}{\spaceskip=0pt\relax}
\providecommand{\BIBentryALTinterwordstretchfactor}{4}
\providecommand{\BIBentryALTinterwordspacing}{\spaceskip=\fontdimen2\font plus
\BIBentryALTinterwordstretchfactor\fontdimen3\font minus \fontdimen4\font\relax}
\providecommand{\BIBforeignlanguage}[2]{{%
\expandafter\ifx\csname l@#1\endcsname\relax
\typeout{** WARNING: IEEEtran.bst: No hyphenation pattern has been}%
\typeout{** loaded for the language `#1'. Using the pattern for}%
\typeout{** the default language instead.}%
\else
\language=\csname l@#1\endcsname
\fi
#2}}
\providecommand{\BIBdecl}{\relax}
\BIBdecl

\bibitem{Janner22:DiffPlanning}
M.~Janner, Y.~Du, J.~Tenenbaum, and S.~Levine, ``Planning with diffusion for flexible behavior synthesis,'' in \emph{Intl. Conf. on Machine Learning ({ICML})}, 2022, pp. 9902--9915.

\bibitem{Reuss23:GoalCondDiff}
M.~Reuss, M.~Li, X.~Jia, and R.~Lioutikov, ``Goal-conditioned imitation learning using score-based diffusion policies,'' in \emph{Robotics: Science and Systems ({R:SS})}, 2023.

\bibitem{Wang23:DiffPolORL}
Z.~Wang, J.~J. Hunt, and M.~Zhou, ``Diffusion policies as an expressive policy class for offline reinforcement learning,'' in \emph{Intl. Conf. on Learning Representations ({ICLR})}, 2023.

\bibitem{Cheng23:DiffPolicy}
C.~Chi, S.~Feng, Y.~Du, Z.~Xu, E.~Cousineau, B.~Burchfiel, and S.~Song, ``Diffusion policy: Visuomotor policy learning via action diffusion,'' in \emph{Robotics: Science and Systems ({R:SS})}, 2023.

\bibitem{luo22:DiffModels}
C.~Luo, ``Understanding diffusion models: A unified perspective,'' \emph{arXiv preprint arXiv2208.11970}, 2022.

\bibitem{Yang23:DiffModelsSurvey}
L.~Yang, Z.~Zhang, Y.~Song, S.~Hong, R.~Xu, Y.~Zhao, W.~Zhang, B.~Cui, and M.-H. Yang, ``Diffusion models: A comprehensive survey of methods and applications,'' \emph{ACM Comput. Surv.}, vol.~56, no.~4, 2023.

\bibitem{Huang22:RiemannianDM}
C.-W. Huang, M.~Aghajohari, J.~Bose, P.~Panangaden, and A.~Courville, ``Riemannian diffusion models,'' in \emph{Neural Information Processing Systems ({NeurIPS})}, 2022.

\bibitem{Lipman23:FlowMatching}
Y.~Lipman, R.~T.~Q. Chen, H.~Ben-Hamu, M.~Nickel, and M.~Le, ``Flow matching for generative modeling,'' in \emph{Intl. Conf. on Learning Representations ({ICLR})}, 2023.

\bibitem{Davtyan23:FMvideo}
A.~Davtyan, S.~Sameni, and P.~Favaro, ``Efficient video prediction via sparsely conditioned flow matching,'' in \emph{Intl. Conf. on Computer Vision ({ICCV})}, 2023, pp. 23\,206--23\,217.

\bibitem{Liu24:FMspeech}
A.~H. Liu, M.~Le, A.~Vyas, B.~Shi, A.~Tjandra, and W.-N. Hsu, ``Generative pre-training for speech with flow matching,'' in \emph{Intl. Conf. on Learning Representations ({ICLR})}, 2024.

\bibitem{Bose24:FMproteinSE3}
J.~Bose, T.~Akhound-Sadegh, K.~FATRAS, G.~Huguet, J.~Rector-Brooks, C.-H. Liu, A.~C. Nica, M.~Korablyov, M.~M. Bronstein, and A.~Tong, ``{SE}(3)-stochastic flow matching for protein backbone generation,'' in \emph{Intl. Conf. on Learning Representations ({ICLR})}, 2024.

\bibitem{Lemme2015:LasaDataset}
A.~Lemme, Y.~Meirovitch, M.~Khansari-Zadeh, T.~Flash, A.~Billard, and J.~J. Steil, ``Open-source benchmarking for learned reaching motion generation in robotics,'' \emph{Paladyn, Journal of Behavioral Robotics}, vol.~6, no.~1, 2015.

\bibitem{Papamakarios21:NormalizingFlows}
G.~Papamakarios, E.~Nalisnick, D.~J. Rezende, S.~Mohamed, and B.~Lakshminarayanan, ``Normalizing flows for probabilistic modeling and inference,'' \emph{Journal of Machine Learning Research}, vol.~22, no.~57, pp. 1--64, 2021.

\bibitem{Rana2020:EuclideanizingFlows}
M.~A. Rana, A.~Li, D.~Fox, B.~Boots, F.~Ramos, and N.~Ratliff, ``Euclideanizing flows: Diffeomorphic reduction for learning stable dynamical systems,'' in \emph{Conference on Learning for Dynamics and Control (L4DC)}, 2020, pp. 630--639.

\bibitem{Khader21:StableNF}
S.~A. Khader, H.~Yin, P.~Falco, and D.~Kragic, ``Learning stable normalizing-flow control for robotic manipulation,'' in \emph{{IEEE} Intl. Conf. on Robotics and Automation ({ICRA})}, 2021, pp. 1644--1650.

\bibitem{Urain20:ImitationFlow}
J.~Urain, M.~Ginesi, D.~Tateo, and J.~Peters, ``Imitationflow: Learning deep stable stochastic dynamic systems by normalizing flows,'' in \emph{{IEEE/RSJ} Intl. Conf. on Intelligent Robots and Systems ({IROS})}, 2020, pp. 5231--5237.

\bibitem{Urain22:StableSE3}
J.~Urain, D.~Tateo, and J.~Peters, ``Learning stable vector fields on lie groups,'' \emph{IEEE Robotics and Automation Letters}, vol.~7, no.~4, pp. 12\,569--12\,576, 2022.

\bibitem{Zhang23:RiemannianStableDS}
J.~Zhang, H.~B. Mohammadi, and L.~Rozo, ``Learning {R}iemannian stable dynamical systems via diffeomorphisms,'' in \emph{Conference on Robot Learning ({CoRL})}, 2023, pp. 1211--1221.

\bibitem{Chen24:RiemannianFM}
R.~T.~Q. Chen and Y.~Lipman, ``Flow matching on general geometries,'' in \emph{Intl. Conf. on Learning Representations ({ICLR})}, 2024.

\bibitem{DoCarmo92:RiemannianGeometry}
M.~do~Carmo, \emph{{R}iemannian Geometry}.\hskip 1em plus 0.5em minus 0.4em\relax Birkh\"auser Basel, 1992.

\bibitem{Lee18:RiemannianManifolds}
J.~M. Lee, \emph{Introduction to {R}iemannian Manifolds}.\hskip 1em plus 0.5em minus 0.4em\relax Springer, 2018.

\bibitem{Tong23:CFMoptimaltransport}
A.~Tong, K.~Fatras, N.~Malkin, G.~Huguet, Y.~Zhang, J.~Rector-Brooks, G.~Wolf, and Y.~Bengio, ``Improving and generalizing flow-based generative models with minibatch optimal transport,'' \emph{Transactions on Machine Learning Research (TMLR)}, 2024.

\bibitem{DormandPrince80:DOPRI}
J.~Dormand and P.~Prince, ``A family of embedded runge-kutta formulae,'' \emph{Journal of Computational and Applied Mathematics}, vol.~6, no.~1, pp. 19--26, 1980.

\bibitem{politorchdyn}
M.~Poli, S.~Massaroli, A.~Yamashita, H.~Asama, J.~Park, and S.~Ermon, ``{TorchDyn}: Implicit models and neural numerical methods in {PyTorch},'' \emph{arXiv preprint arXiv:2009.09346}, 2020.

\bibitem{Ramachandran17:Swish}
P.~Ramachandran, B.~Zoph, and Q.~V. Le, ``Searching for activation functions,'' \emph{arXiv preprint arXiv:1710.05941}, 2017.

\bibitem{Polyak92:MovingAverage}
B.~T. Polyak and A.~B. Juditsky, ``Acceleration of stochastic approximation by averaging,'' \emph{SIAM Journal on Control and Optimization}, vol.~30, no.~4, pp. 838--855, 1992.

\bibitem{Mardia99:DirectionalStats}
K.~V. Mardia and P.~E. Jupp, \emph{Distributions on Spheres}.\hskip 1em plus 0.5em minus 0.4em\relax John Wiley and Sons, Ltd, 1999, ch.~9, pp. 159--192.

\bibitem{Galaz-Garcia22:WrappedHomogenous}
F.~Galaz-Garcia, M.~Papamichalis, K.~Turnbull, S.~Lunagomez, and E.~Airoldi, ``Wrapped distributions on homogeneous {R}iemannian manifolds,'' \emph{arXiv preprint 2204.09790}, 2022.

\bibitem{Nichol21:ImprovedDenoisingDPM}
A.~Q. Nichol and P.~Dhariwal, ``Improved denoising diffusion probabilistic models,'' in \emph{Intl. Conf. on Machine Learning ({ICML})}, ser. Proceedings of Machine Learning Research, vol. 139, 2021, pp. 8162--8171.

\bibitem{Yoon23:DiffusionMemorization}
T.~Yoon, J.~Y. Choi, S.~Kwon, and E.~K. Ryu, ``Diffusion probabilistic models generalize when they fail to memorize,'' in \emph{ICML 2023 Workshop on Structured Probabilistic Inference {\&} Generative Modeling}, 2023.

\bibitem{Balasubramanian15:Smoothness}
S.~Balasubramanian, A.~Melendez-Calderon, A.~Roby-Brami, and E.~Burdet, ``On the analysis of movement smoothness,'' \emph{Journal of NeuroEngineering and Rehabilitation}, vol.~12, no. 112, 2015.

\bibitem{Jaquier24:Fallacy}
N.~Jaquier, L.~Rozo, and T.~Asfour, ``Unraveling the single tangent space fallacy: An analysis and clarification for applying {R}iemannian geometry in robot learning,'' in \emph{{IEEE} Intl. Conf. on Robotics and Automation ({ICRA})}, 2024, pp. 242--249.

\bibitem{Lou23:ScalingRiemannianDM}
A.~Lou, M.~Xu, A.~Farris, and S.~Ermon, ``Scaling {R}iemannian diffusion models,'' in \emph{Neural Information Processing Systems ({NeurIPS})}, 2023.

\bibitem{Mandlekar22:LearningFromOfflineDemos}
A.~Mandlekar, D.~Xu, J.~Wong, S.~Nasiriany, C.~Wang, R.~Kulkarni, L.~Fei-Fei, S.~Savarese, Y.~Zhu, and R.~Mart\'in-Mart\'in, ``What matters in learning from offline human demonstrations for robot manipulation,'' in \emph{Conference on Robot Learning ({CoRL})}, ser. Proceedings of Machine Learning Research, vol. 164, 2022, pp. 1678--1690.

\bibitem{Wu18:GroupNormalization}
Y.~Wu and K.~He, ``Group normalization,'' in \emph{European conference on computer vision ({ECCV})}, 2018, pp. 3--19.

\end{thebibliography}

\end{document}